\documentclass[10pt,twocolumn,letterpaper]{article}

\usepackage{cvpr}              
\usepackage{multirow}
\usepackage{pifont}
\usepackage{graphicx}
\definecolor{cvprblue}{rgb}{0.21,0.49,0.74}
\usepackage[pagebackref,breaklinks,colorlinks,allcolors=cvprblue]{hyperref}

\def\paperID{*****} 

\title{Your VLM Already Knows When: Training-Free Temporal Grounding by Asking Yes or No}

\author{Ji Huang\\
Queen's University Belfast\\
{\tt\small jhuang28@qub.ac.uk}
\and
Barry Devereux\\
Queen's University Belfast\\
{\tt\small B.Devereux@qub.ac.uk}
\and
Hui Wang\\
Queen's University Belfast\\
{\tt\small h.wang@qub.ac.uk}
}

\begin{document}
\maketitle

\begin{abstract}
Multimodal LLMs that recognise events reliably still fail to say when
they happen. Prompted for timestamps, strong VLMs reach as little as
$3.8\%$ R@0.5 on Charades-STA, and $77$ to $80\%$ of their wrong
predictions carry low output entropy: the models are confidently wrong,
and entropy-based error detection stays below a random classifier.
We show that this failure lives in the task interface, not in perception.
Holding the weights fixed, replacing timestamp regression with a
coarse-to-fine scan of binary questions, whose first-token probabilities
are consumed only as a ranking, raises R@0.5 by $28$ to $50$ points
across four frozen backbones.
The residual failures decompose into two measurable axes: a perception
axis that moves with the backbone, and a geometry axis that is
analytically predictable from the ratio of the output-window and event
widths.
FV-Action, the training-free method built on this analysis, reaches
$56.8\%$ R@0.5 on Charades-STA, above the same backbone's native
grounding pipeline and the strongest training-free result on this
benchmark; it surpasses every TVG-trained model evaluated zero-shot on
TACoS, and improves over direct prediction on ActivityNet Captions and
QVHighlights, with no temporal supervision at any stage.
\end{abstract}

\section{Introduction}
\label{sec:intro}
\begin{figure}[t]
  \centering
  \includegraphics[width=\linewidth]{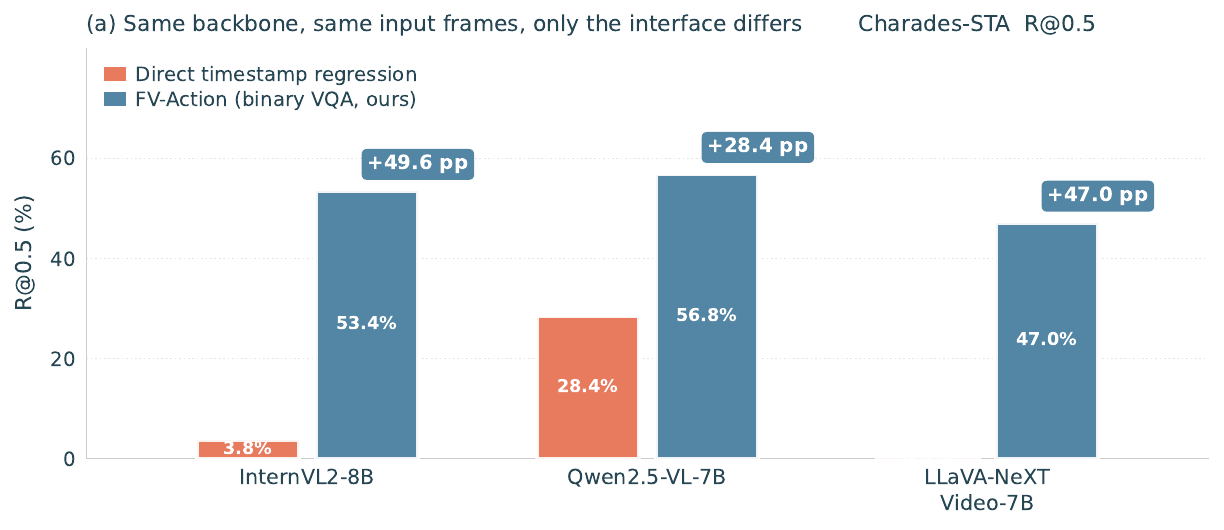}
  \caption{
    \textbf{The interface, not perception, is the bottleneck.}
    Three frozen backbones under direct timestamp prediction versus the
    same weights under FV-Action's binary scanning (Charades-STA R@0.5).
    Nothing about the models changes; only the way the task is posed and
    read out does.
  }
  \label{fig:teaser}
\end{figure}

Temporal video grounding (TVG) asks where in an untrimmed video the event
described by a natural-language query
occurs~\cite{gao2017tall,krishna2017dense,regneri2013grounding}.
The task demands joint reasoning over visual content and language,
precisely what multimodal large language models (MLLMs) are built
for~\cite{chen2024expanding,bai2025qwen25vltechnicalreport}, and the natural way to use one is
to ask it for timestamps directly.
This fails, and not marginally.
InternVL2-8B~\cite{chen2024expanding}, strong across diverse visual
benchmarks, reaches $3.76\%$ R@0.5 on Charades-STA~\cite{gao2017tall} when
prompted for timestamps.
Models fine-tuned specifically for grounding fare little better
(TimeChat~\cite{ren2024timechat} $31.2\%$, LITA~\cite{huang2024lita}
$8.87\%$), and the strongest general-purpose VLM under an optimised native
temporal pipeline reaches $53.6\%$~\cite{zheng2026omnivtg}.
The same pattern holds on ActivityNet Captions and TACoS.
Why does direct timestamp prediction fail so systematically for models
that excel at visual reasoning?

A natural hypothesis is that the models are guessing, in which case their
own uncertainty should expose the errors.
The opposite is true.
Across TimeChat, LITA, and VTG-LLM, $77$ to $80\%$ of wrong predictions
carry low output entropy: the models are \emph{confidently} wrong.
Entropy-based error detection reaches F1 below $0.36$ on all three, under
the $0.50$ of a random classifier, so no amount of confidence
thresholding separates right from wrong.
The failure is systematic rather than epistemic, which points away from
what the model knows and toward how it is asked to answer.

We test that reading directly.
For each sample we compare the model's Yes-probability on the ground-truth
window against an equal-width, non-overlapping window from the same video,
using the binary question \emph{``Does this clip show \{query\}?''}, and
summarise the outcome by the signal gap
$\Delta = P(\text{Yes}\mid\text{GT}) - P(\text{Yes}\mid\text{neg})$ and
the rate at which the ground-truth clip wins.
On Charades-STA perception is intact ($\Delta = 0.308$; the true clip wins
$80.5\%$ of pairs), and holding the model fixed while replacing timestamp
regression with a coarse-to-fine scan over these questions
(Sec.~\ref{sec:method}) raises R@0.5 from $3.76$ to $53.39$ for
InternVL2-8B and from $28.41$ to $56.77$ for Qwen2.5-VL-7B.
Where the model perceives the event, the interface, not perception, is
the bottleneck.

The failures that remain decompose along two measurable and separable
axes, and this decomposition organises the paper.
The \emph{perception} axis is a property of the backbone.
On TACoS the per-clip signal gap rises from $0.186$ (InternVL2) to
$0.260$ (Qwen2.5-VL) on identical samples ($p{=}0.0009$), and deployed
accuracy rises with it, which makes the gap a cheap pre-deployment
diagnostic of which model will scan better.
The \emph{geometry} axis is a property of the output form.
With a fixed window of width $W$, overlap with an event of width $g$
cannot exceed $\min(W,g)/\max(W,g)$, so accuracy at a given threshold is
predictable before any model runs; on ActivityNet the ground-truth
buckets outside the feasible band collapse exactly as the bound dictates,
even though the located peak still falls inside the event on up to
$95\%$ of them.
The two axes prescribe different remedies, a better instrument for one
and a better output rule for the other, and neither is repaired by
processing the scores in between: smoothing, contrastive correction,
denser scanning, and prompt rewriting all leave the located peak where
it was (Sec.~\ref{sec:analysis}).

The mechanism behind the interface gap is a format mismatch.
Decimal timestamps conditioned on a language query are largely absent
from pre-training corpora, so a model asked to emit them falls back on
distributional priors and produces plausible values with high confidence,
which is exactly the low-entropy failure observed above.
Binary VQA sidesteps the missing format, and scanning consumes only the
\emph{ranking} of its answers across clips.
Ranking is a weaker requirement than calibration, and it is the one the
models meet: scores that look poorly calibrated in isolation still order
clips reliably within a video (Sec.~\ref{sec:method}).

\textbf{FV-Action} operationalises this reading with deliberate
simplicity.
It poses one binary question per clip over a coarse-to-fine hierarchy of
video segments, selects the peak by rank, and emits a fixed-width window;
the backbone stays frozen and performs only the judgement it is
demonstrably good at.
On Charades-STA it reaches $56.8\%$ R@0.5, surpassing the \emph{same}
backbone's optimised native pipeline ($53.6\%$~\cite{zheng2026omnivtg})
on identical frozen weights, and doing so on the stricter R@0.7 as well
($29.7$ vs.\ $28.5$).
The gain is a property of the interface, not the model: it holds across
four backbones with improvements of $28$ to $50$ points, including
InternVL2-8B from ${\sim}3.8\%$ to $53.4\%$.
Without any temporal training, FV-Action remains competitive with systems
that pay for it, whether by full supervision
(UniVTG~\cite{lin2023univtg}, $60.2\%$), reinforcement learning
(Time-R1~\cite{wang2026time}, $60.8\%$), or large-scale annotation
(OmniVTG~\cite{zheng2026omnivtg}, $63.2\%$ with $359$K pairs), and it
improves over direct prediction on four benchmarks
(Charades-STA, ActivityNet Captions~\cite{krishna2017dense},
TACoS~\cite{regneri2013grounding}, and
QVHighlights~\cite{lei2021detecting}).

\noindent Our contributions are as follows:
\begin{itemize}
\item A diagnostic study showing that \textbf{77--80\% of MLLM temporal
      hallucinations are confident (low-entropy)}, with entropy-based error
      detection achieving F1 $<$ 0.36 across TimeChat, LITA, and VTG-LLM,
      below the random-classifier baseline of 0.50.
      This establishes that the dominant failure mode is systematic format
      hallucination, not epistemic uncertainty.
\item A controlled same-backbone comparison isolating the \textbf{interface}
      as the bottleneck. With weights, frames, and prompting held fixed,
      replacing timestamp regression with binary VQA scanning raises R@0.5 by
      $28$ to $50$ points on Charades-STA across four frozen backbones, and
      lifts Qwen2.5-VL-7B above its own native grounding pipeline on both
      R@0.5 and R@0.7.
\item An analysis decomposing the residual failures into two measurable axes.
      The \textbf{perception} axis moves with the backbone. On TACoS the
      per-clip oracle gap rises from $0.186$ to $0.260$ when InternVL2 is
      replaced by Qwen2.5-VL, a paired difference on identical samples
      ($p{=}0.0009$), and deployed accuracy rises with it. The
      \textbf{geometry} axis is governed by the fit between the output window
      and the event width, and is analytically predictable from their ratio
      alone.
\item \textbf{FV-Action}, a training-free grounding framework built on this
      analysis. It poses only binary questions to a frozen VLM and reads a
      ranking rather than a calibrated score, reaching $56.8\%$ R@0.5 on
      Charades-STA, the strongest training-free result on this benchmark,
      with consistent gains over direct prediction on ActivityNet Captions
      and TACoS.
\end{itemize}

%

\section{Related Work}
\label{sec:related}

\paragraph{Supervised video temporal grounding.}
Video temporal grounding (TVG) localises a natural-language query in a video as a
$[s,e]$ interval~\cite{gao2017tall}.
Span-based methods enumerate candidate moments and rank them with cross-modal
similarity (2D-TAN~\cite{zhang2020learning}) or contrastive metric learning
(MMN~\cite{wang2022negative}).
Proposal-free approaches directly regress start/end offsets from dense
feature interactions (UniVTG~\cite{lin2023univtg}), achieving over $60\%$ R@0.5
on Charades-STA with full supervision.
Despite strong in-domain performance, these models require complete TVG annotations
and generalise poorly across domains without retraining.
Our method requires no TVG annotations at any stage.

\paragraph{Multimodal LLMs for TVG.}
Integrating temporal grounding into video LLMs has been pursued along several lines.
LITA~\cite{huang2024lita} and VTG-LLM~\cite{guo2025vtg} insert special timestamp
tokens into the vocabulary so the model can emit temporal boundaries as part of its
output sequence.
TimeChat~\cite{ren2024timechat}, VTimeLLM~\cite{huang2024vtimellm},
ChatVTG~\cite{qu2024chatvtg}, Momentor~\cite{qian2024momentor},
and HawkEye~\cite{wang2024hawkeye} instead fine-tune on TVG instruction data,
while Enrich-and-Detect~\cite{pramanick2025enrich} combines a VLM with an external
proposal network.
At larger training scales, TimeMarker~\cite{chen2024timemarker},
TRACE~\cite{guo2024trace}, TimeSuite~\cite{zeng2025timesuite},
and UniTime~\cite{li2026universal} incorporate temporal annotations during
pre-training to strengthen localisation.
Despite these efforts, most models fall short of fully supervised specialists,
partly because generating sub-second timestamps forces a vocabulary trained on
natural language to operate as a numerical regressor.
We show that this is a formulation bottleneck rather than a capacity
bottleneck, in TVG-tuned and general-purpose
models~\cite{chen2024expanding,bai2025qwen25vltechnicalreport,ye2025mplug}
alike: InternVL2-8B predicts timestamps at $3.76\%$ R@0.5 yet discriminates
clips via binary VQA at AUROC $0.782$, and switching the interface recovers
$28$ to $50$ pp across four frozen backbones (Table~\ref{tab:diagnostic}).
Time-R1~\cite{wang2026time} addresses the regression difficulty with
reinforcement learning on 2,500 TVG samples, reaching $60.8\%$ R@0.5.
Concurrently, OmniVTG~\cite{zheng2026omnivtg} independently identifies the same
formulation bottleneck and proposes a three-stage training pipeline (supervised
fine-tuning, chain-of-thought fine-tuning, and GRPO reinforcement learning) on a
proprietary dataset of 359{,}K temporal annotations spanning 2,124 hours, reaching
$63.2\%$ R@0.5 on Charades-STA.
FV-Action achieves $56.8\%$ with no training on TVG data, within striking
distance of OmniVTG, showing that interface redesign alone captures the majority
of the gain.

\paragraph{Zero-shot and training-free TVG.}
LLM4VG~\cite{feng2023llm4vg} generates per-second captions and feeds them to an LLM
for temporal reasoning, reaching $11.8\%$ R@0.5 on Charades-STA, only marginally
above a random baseline, because language-space reasoning cannot substitute for
direct visual evidence.
Moment-GPT~\cite{xu2025zero} assembles a multi-model pipeline combining
language rewriting, video captioning, and LLM reasoning, reaching $38.4\%$,
while GranAlign~\cite{jeon2026granalign} aligns visual granularities via CLIP
matching and achieves $39.6\%$, and BTDP~\cite{deng2025boundary} reaches
$40.0\%$ with dense captioning and LLM alignment.
FV-Action differs from all of these in that it directly queries a single VLM with
binary visual questions at each candidate clip, requiring no caption generation and
no auxiliary models.
This reformulation raises the training-free Charades-STA R@0.5 to $56.8\%$,
a $16.8$ pp improvement over the prior best ($40.0\%$).

\paragraph{Uncertainty as an error signal.}
A parallel line asks whether a model's own uncertainty exposes its errors.
Semantic entropy detects confabulations in free-form text
generation~\cite{kuhn2023semantic}, and binary
interrogation probes such as POPE~\cite{li2023evaluating} audit object
hallucination in VLMs with yes-or-no questions.

Our diagnostic study carries this question to temporal grounding and returns
a negative answer: across three TVG-tuned models, $77$ to $80\%$ of wrong
predictions carry low output entropy, and entropy-based error detection
stays below a random classifier (F1 $< 0.36$).
The failure is systematic rather than epistemic, so no confidence signal can
gate it away; FV-Action instead replaces the unreliable self-report with
direct visual evidence, the ranking of binary judgements over clips, which
is itself a POPE-style interrogation turned into a localisation instrument.

\section{Method}
\label{sec:method}

\subsection{Problem Formulation}
\label{sec:formulation}

Given an untrimmed video $V$ of duration $T$ seconds and a natural-language
query $q$, temporal video grounding (TVG) seeks the interval
$[s, e] \subset [0, T]$ during which the described event occurs.
Two interfaces can extract this interval from a multimodal LLM.
Under the \emph{generative} interface, the model is prompted to emit decimal
timestamps directly.
Under the \emph{discriminative} interface, the model answers only a binary
VQA question about a short clip, and localisation is performed by an
external scan over clip positions.
The same frozen backbone behaves very differently under the two.
InternVL2-8B reaches 3.76 R@0.5 when generating timestamps, yet separates
clips inside the ground-truth interval from equal-width non-overlapping
negatives at a discrimination rate of 80.5\% (Sec.~\ref{sec:analysis});
switching only the interface raises its R@0.5 to 53.39.
Because the backbone is identical under the two interfaces and only the
output format changes, this gap isolates the interface as the cause: the
grounding ability is present but expressed through the wrong channel.
The mechanism is a format mismatch. Decimal timestamps conditioned on a
language query are largely absent from pre-training corpora, so the model
falls back on language priors and produces plausible but ungrounded
values, which is why these errors carry low entropy (Sec.~\ref{sec:intro}).
FV-Action is the discriminative interface made operational. It introduces
no new model and reroutes the task through the judgement the backbone
already performs reliably.

\subsection{Framework Overview}
\label{sec:overview}


The pipeline has three stages.
A coarse scan places $K_c$ clip centres uniformly over $[0, T]$ and asks
the frozen backbone one binary question per clip, yielding an
action-probability curve.
Non-maximum suppression keeps the top $N_p$ candidate peaks, a fine scan
re-examines the neighbourhood of each candidate with $K_f$ clips, and a
joint score over both scans selects the final peak $t^{*}$.
The prediction is a fixed-width window centred on $t^{*}$.
The decomposition is not plumbing but the operational form of the
analysis in Sec.~\ref{sec:analysis}: ask the model only the question it
answers reliably, consume only the ranking of its answers, and emit only
the geometry the signal quality supports. Each stage exists because the
corresponding shortcut fails, and Sec.~\ref{sec:rationale} traces each
choice to the failure it answers.
The framework involves no training, no fine-tuning, and no temporal
supervision; the only learned component is the frozen backbone itself.

\subsection{Stage 1: Coarse Scanning with a Frozen Measurement Layer}
\label{sec:stage1}

\noindent\textbf{Sampling.}
We place $K_c$ clip centres $\{t_i\}$ uniformly over $[0, T]$.
Each clip is represented by $F{=}5$ frames extracted symmetrically about
$t_i$ and spaced $\delta$ apart, where $\delta$ is one third of the probe
spacing, so a probe spans $4\delta$; frames are resized to
$448{\times}448$.
Both settings sit at a measured optimum. Fewer frames or a narrower span
lose accuracy, more frames over the same span add nothing, and a wider span
hurts (Sec.~\ref{sec:ablation}).
The budget $K_c$ scales with video duration so that the spacing between
probes stays within a few seconds: $K_c{=}12$ on Charades-STA
(about 30\,s videos, 2.5\,s spacing), $K_c{=}16$ on ActivityNet Captions
(125\,s, 7.8\,s), $K_c{=}16$ on QVHighlights (150\,s, 9.4\,s), and
$K_c{=}96$ on TACoS (400\,s, 4.2\,s).
This spacing rule is also where accuracy peaks: on Charades-STA both a
sparser and a denser scan are worse than the deployed budget
(Sec.~\ref{sec:ablation}), so the scan cannot be improved by sampling the
timeline harder; what does improve it is enriching the evidence each probe
sees, which is the role of the five-frame probe.
Each clip receives the fixed prompt
\texttt{"Does this clip show \{$\tilde{q}$\}? Answer Yes or No."},
where $\tilde{q}$ is the query with trailing punctuation removed.
This single template is used unchanged for every clip, query, backbone and
dataset; no per-corpus prompt engineering is involved.

\noindent\textbf{Probability readout.}
From the logits of the first generated token we compute
\begin{equation}
  P(\text{Yes} \mid t_i) =
  \frac{\displaystyle\sum_{v \in \mathcal{V}_\text{yes}} p_v}
       {\displaystyle\sum_{v \in \mathcal{V}_\text{yes}} p_v
        + \sum_{v \in \mathcal{V}_\text{no}}  p_v},
  \label{eq:prob}
\end{equation}
where $\mathcal{V}_\text{yes}$ and $\mathcal{V}_\text{no}$ are the token
sets covering the surface forms of \texttt{Yes} and \texttt{No}.
No decoding beyond the first token is required.
Reading position zero is safe rather than merely convenient: probing $3{,}600$
clips under the deployed configuration, the mass
$\sum_{\mathcal{V}_\text{yes}} p_v + \sum_{\mathcal{V}_\text{no}} p_v$ that
Eq.~\ref{eq:prob} normalises by never falls below $0.998$, the first-token
argmax is a \texttt{Yes}/\texttt{No} surface form on every clip without
exception, and greedy decoding never places its answer at a later position.
The denominator is therefore a normalisation over essentially all of the
first-token distribution, not a renormalisation of a small residue.
Downstream stages consume only the \emph{ranking} of these scores across
clips, never their calibrated values, so a backbone whose absolute
probabilities are poorly calibrated can still be a strong scanner.
The same probe makes the distinction concrete. The discrete answer is
\texttt{No} on $88.3\%$ of clips even though $27.9\%$ of them fall inside the
ground-truth interval: the median score on a positive clip is $0.202$ against
$0.042$ on a negative one, so the two populations are well ordered
(AUROC $0.732$) while both sit far below the natural decision boundary, which
recovers only $26.4\%$ of positive clips at $F_1{=}0.368$.
Retuning the threshold does not repair this. The optimum lies at $0.10$ rather
than $0.5$ and still reaches only $F_1{=}0.536$, because the score scale drifts
across videos: the tenth and ninetieth percentiles of the per-video maximum are
$0.165$ and $0.906$, so any global threshold is simultaneously too strict for
one video and too permissive for another. Ranking is a within-video operation
and is immune to that drift; taking the per-video argmax with no threshold at
all lands inside the ground-truth interval on $66.0\%$ of videos.
A hard \texttt{Yes}/\texttt{No} readout would discard three quarters of the
positive evidence and leave the scan curve nearly flat, while the continuous
score preserves exactly the ordering that this conservatism destroys. The
interface change therefore requires probabilities and not answers.
Sec.~\ref{sec:analysis} examines this distinction directly.

\noindent\textbf{A pluggable measurement layer.}
Stage 1 treats the backbone as a measurement instrument. Nothing in the
pipeline depends on its architecture, so the backbone can be exchanged
freely. We instantiate it with InternVL2-8B~\cite{chen2024internvl},
Qwen2.5-VL-7B~\cite{bai2025qwen25vltechnicalreport}, LLaVA-NeXT-Video-7B~\cite{zhang2024llavanextvideo},
and mPLUG-Owl3-7B~\cite{ye2025mplug}, all frozen.
Because per-clip discriminability is one of the two failure axes we
identify, upgrading the instrument converts directly to accuracy where
that axis binds: on TACoS, replacing InternVL2 with Qwen2.5-VL raises
R@0.5 from 8.6 to 13.0 with the pipeline unchanged
(Sec.~\ref{sec:analysis}).

\subsection{Stage 2: Peak Localization}
\label{sec:stage2}

The coarse curve typically contains several local maxima.
We keep the top $N_p$ of them under non-maximum suppression with radius
$r_f$: candidates are selected greedily by score, and each selection
suppresses all positions within $r_f$ seconds,
\begin{equation}
  \hat t_{j} = \arg\max_{t_i}\;
  P(\text{Yes} \mid t_i)
  \quad \text{s.t.} \quad
  |t_i - \hat t_{j'}| > r_f \;\; \forall j' < j.
  \label{eq:nms}
\end{equation}
We fix $N_p{=}3$ for every corpus. The sweep in Sec.~\ref{sec:ablation}
shows the multi-candidate gain saturates by the second or third candidate on
every corpus and is flat thereafter, so a single shared value suffices and no
per-corpus tuning is needed.
Around each candidate $\hat t_j$ a fine scan places $K_f{=}8$ clips
uniformly within $[\hat t_j - r_f,\, \hat t_j + r_f]$, giving scores
$\{P^{f}_{jk}\}$, with $r_f{=}8$\,s, $25$\,s, $25$\,s, and $30$\,s on
Charades-STA, ActivityNet, QVHighlights, and TACoS respectively, matching
each corpus's event scale.
Candidates are then ranked by the joint score
\begin{equation}
  s_j = P(\text{Yes} \mid \hat t_j)^{\,\alpha}
        \cdot \Bigl(\max_{k}\,P^{f}_{jk}\Bigr)^{\!1-\alpha},
  \qquad \alpha = 0.5,
  \label{eq:combined}
\end{equation}
a geometric mean that weighs the two scan resolutions equally and is
insensitive to their absolute scales. We fix $\alpha{=}0.5$ rather than
learning it, keeping the pipeline free of trained parameters;
Sec.~\ref{sec:ablation} shows accuracy is stable across $\alpha$.
The final peak $t^{*}$ is the fine-scan argmax under the winning
candidate.

\subsection{Stage 3: Window Generation and Its Geometric Characterisation}
\label{sec:stage3}

\noindent\textbf{Output rule and the width setting.}
The prediction is a fixed-width window centred on the peak and truncated
at the video boundary,
\begin{equation}
  [\hat s, \hat e] =
  \bigl[\max(0,\; t^{*} \!-\! W/2),\;
        \min(T,\; t^{*} \!+\! W/2)\bigr].
  \label{eq:window}
\end{equation}
Here $W$ is a per-corpus expectation of event scale ($8$\,s on
Charades-STA, $20$\,s on TACoS, $80$\,s on ActivityNet Captions), and the
one setting in the pipeline with the character of a domain prior, so we
state exactly where each value comes from. On Charades-STA the deployed
value is the corpus mean event length read from the training split
($7.8$\,s rounded to $8$), and on QVHighlights likewise ($25.1$\,s
rounded to $25$). On the two long-video corpora no single
training statistic reproduces the deployed values, because their event
distributions span two orders of magnitude and any single width is a
compromise; the values sit on the broad plateau that the sensitivity
sweep in Sec.~\ref{sec:ablation} makes explicit, and the same sweep
reports the cost of insisting on the strictly prior-derived alternative.
As $W$ grows, R@0.3 rises and R@0.7 falls while R@0.5 stays within about
one point across a wide band, exactly the trade that
$\mathrm{IoU}=\min(W,g)/\max(W,g)$ predicts, so a choice within the
plateau shifts error between thresholds rather than manufacturing
accuracy. The remaining settings ($K_c$, $r_f$, $N_p$) follow the video's
temporal resolution and are not tuned to the target.

\noindent\textbf{Geometric characterisation.}
The fixed window makes the attainable accuracy analytically predictable.
If the peak is perfectly centred on an event of width $g$, the overlap is
\begin{equation}
  \mathrm{IoU}(W, g) = \frac{\min(W, g)}{\max(W, g)},
  \label{eq:geo}
\end{equation}
so $\mathrm{IoU} \geq 0.5$ requires $g \in [\,W/2,\; 2W\,]$, and outside
this band R@0.5 is zero regardless of how well the peak is localised.
This closed form is a testable prediction, not a disclaimer: on
ActivityNet the ground-truth buckets lying outside the feasible band show
exactly zero R@0.5, matching the bound. Where the band is satisfied,
final accuracy sits close to this geometric ceiling, indicating that
post-peak error is small and that performance is governed primarily by
window--event geometry (Sec.~\ref{sec:analysis}).

\noindent\textbf{Adaptive width as a documented variant.}
A natural alternative sets $W$ per sample from the full width at half
maximum of the coarse curve, clipped to a global physical guardrail of
$[2, 120]$\,s whose bounds reject single-probe noise spikes and prevent
divergence on flat curves, and which contains no dataset statistics.
This variant recovers long events that no single fixed width can cover
but degrades short homogeneous events, a two-sided pattern that is stable
under both an absolute and a baseline-corrected half-maximum definition.
Sec.~\ref{sec:ablation} quantifies both directions.
A gating rule derived from the scan curve cannot separate the two
regimes: the curve's full width at half maximum is uncorrelated with
event duration (Sec.~\ref{sec:analysis}), so no global threshold on it
recovers the adaptive gain without reintroducing the fixed-window loss.
We therefore keep the fixed window as the default.

\subsection{Design Rationale: What the Two-Axis Analysis Dictates}
\label{sec:rationale}

Each design choice above follows from one of the two failure axes we
measure in Sec.~\ref{sec:analysis}.
We sample uniformly rather than adaptively focusing the scan because
sampling is not the bottleneck. Under the budgets above, the analytic
probability that at least one probe lands inside the ground-truth
interval is 0.87 to 0.95 across the three corpora, while the measured
peak hit rate is far lower, placing the loss in per-clip discrimination
rather than in coverage.
We predict a peak plus a window rather than estimating interval
boundaries because boundary estimation demands a signal quality the
per-clip scores do not provide. An interval-estimation variant based on
thresholded score support underperformed peak detection consistently and
is reported as a negative result in Sec.~\ref{sec:analysis}.
Finally, we train nothing. Fine-tuning the backbone would forfeit the
central claim, that the grounding ability already exists in the frozen
model and is buried by the generative interface, and it would forfeit the
property that FV-Action improves for free with every stronger release of
its measurement layer.

\section{Experiments}
\label{sec:experiments}

\subsection{Experimental Setup}
\label{sec:setup}

\paragraph{Datasets.}
We evaluate on three standard TVG benchmarks.
\textbf{Charades-STA}~\cite{gao2017tall} contains 3,720 test samples from indoor activity
videos (mean duration 30.6 s, mean GT length 7.8 s).
\textbf{ActivityNet Captions}~\cite{krishna2017dense} is evaluated on the standard
\texttt{val\_1} split of 17,505 query-video pairs (mean duration 180 s, mean GT
length 37.7 s). As on all prior work, a fraction of the YouTube source videos are
no longer retrievable, so, following common practice, evaluation is restricted to
the videos that remain available; we report on this set without penalising the
method for missing videos, matching the protocol used by the baselines we compare
against.
\textbf{TACoS}~\cite{regneri2013grounding} covers 4,001 test samples from a single cooking
domain (mean video duration 6.7 min), a challenging out-of-domain setting for zero-shot methods.
\textbf{QVHighlights}~\cite{lei2021detecting} contains 1,550 val queries
over 150\,s vlog and news clips (train mean moment length 25.1\,s);
$34\%$ of queries annotate multiple disjoint moments, and a prediction is
scored by its maximum IoU over the annotated windows. As with ActivityNet,
a fraction of source videos ($7\%$) is no longer retrievable, and we
evaluate on the 1,441 available queries.

\paragraph{Metrics.}
Following standard evaluation, we report Recall at IoU threshold $\tau$
($R$@$\tau$, $\tau \in \{0.3, 0.5, 0.7\}$), which measures the fraction of predictions
whose temporal IoU with the ground truth exceeds $\tau$.

\noindent\textbf{Direct regression baseline.}
The direct baseline is held strictly identical across all backbones except
for the frozen vision-language model itself, so that the interface, not the
input protocol, is the only variable.
For every sample we uniformly sample sixteen frames, one at the centre of
each of sixteen equal temporal segments, and present them to the model under
a single verbatim text prompt: the video duration in seconds, the query,
and the instruction to output only the start and end time in seconds.
The prompt states the duration so the model has an explicit temporal anchor;
the interface, not missing time information, is what we test.
The sixteen frames are ingested through each backbone's native video
interface (image tokens for InternVL2, which has no video modality, and the
native video token for Qwen2.5-VL, LLaVA-NeXT-Video, and mPLUG-Owl3), which
is the only component that differs across backbones; no per-frame timestamp
markup is inserted, so no backbone is forced into an unnatural interleaved
image mode.
Generation is greedy with a thirty-two token budget, we parse the first two
decimals in the output as $[\hat s, \hat e]$ clamped to $[0, D]$, and report
the standard temporal IoU, taken as the maximum over ground-truth windows.
A controlled interface comparison requires the two arms to see the video at a
comparable temporal resolution, which we define as the number of distinct
temporal locations sampled across the clip, not the raw frame count. On
Charades-STA the regression arm draws sixteen uniform frames, i.e.\ sixteen
temporal locations, while the FV-Action coarse scan probes $K_c{=}12$ clip
centres, so the two arms cover the timeline at a similar granularity even though
FV-Action reads three frames per clip (thirty-six frames total). If anything the
extra frames give no coverage advantage, since they are clustered at the same
twelve locations. On ActivityNet and TACoS the sixteen-frame regression input
and the sixteen- or ninety-six-probe scan differ in temporal resolution by
design, which would confound the interface with coverage, so we do not report a
controlled regression baseline there.

\paragraph{Implementation details.}
All backbones are used frozen, with no fine-tuning or temporal
supervision at any stage. We evaluate four public checkpoints without
modification: InternVL2-8B ,
Qwen2.5-VL-7B-Instruct,
LLaVA-NeXT-Video-7B, and
mPLUG-Owl3-7B. We run inference in
half precision (\texttt{bfloat16}, or \texttt{float16} for LLaVA-NeXT-Video) on a
single NVIDIA A100; the
full pipeline requires no training and no gradient computation. Frames are
decoded with ffmpeg at $448{\times}448$ and, for each clip, $F{=}3$ frames
are sampled around the clip centre. The binary VQA prompt is fixed across
all datasets and backbones (Sec.~\ref{sec:stage1}); $P(\text{Yes})$ is read
from the first-token logits over the \texttt{Yes}/\texttt{No} surface-form
token sets (Eq.~\ref{eq:prob}). We fix the joint-score exponent
$\alpha{=}0.5$. The three per-corpus settings follow each video's
temporal scale and are summarised in Table~\ref{tab:hparams}; no other
hyperparameter is tuned to the target datasets. For the direct-regression
baseline we generate greedily with a 32-token budget and parse the first
two decimals as $[\hat s,\hat e]$. All reported numbers use a single
inference run (the pipeline is deterministic under greedy decoding).

\paragraph{Inference cost.}
Being training-free trades gradient updates for repeated forward passes at
inference. Each sample requires $K_c + N_p K_f$ binary-VQA forwards: the
$K_c$ coarse probes plus a fine scan of $K_f{=}8$ clips around each of the
$N_p$ surviving candidates. This is 36 forwards on Charades-STA
($12{+}3{\times}8$), 40 on ActivityNet ($16{+}3{\times}8$), and 120 on
TACoS ($96{+}3{\times}8$), against a single forward for direct regression.
Each forward reads only the first-token Yes/No logits (Eq.~\ref{eq:prob}),
with no autoregressive decoding, and the probes are mutually independent, so
they are issued as GPU batches rather than sequentially. On a shared 19.5\,GB
A100 MIG partition the median wall-clock cost per sample is 14.4\,s, 22.4\,s,
and 95.6\,s (Table~\ref{tab:hparams}); on a dedicated A100 the Charades figure
falls to 10.3\,s. The cost is dominated by the backbone's forward pass, not by
video decoding, and scales with the probe budget, so it is the price of
avoiding any training rather than a fixed overhead: it trades a one-time
training and annotation burden, which the reinforcement-learning and
large-scale-SFT baselines pay, for test-time computation that needs no labels.
The budget is also adjustable: reducing $K_c$ trades accuracy for forwards along
a measured curve (Sec.~\ref{sec:ablation}), and the scan inherits any speedup of
the underlying backbone.

\begin{table*}[t]
\centering
\caption{Per-corpus settings. All follow the video temporal scale;
none is tuned on target labels except the exposed width setting $W$
(Sec.~\ref{sec:stage3}). Median latency is measured on a shared A100 MIG
partition; a dedicated A100 is about $1.4\times$ faster.}
\label{tab:hparams}
\begin{tabular}{lccc}
\toprule
 & Charades-STA & ActivityNet & TACoS \\
\midrule
Coarse probes $K_c$      & 12   & 16   & 96   \\
Probe spacing (s)        & 2.5  & 7.8  & 4.2  \\
Fine radius $r_f$ (s)    & 8    & 25   & 30   \\
Fine clips $K_f$         & 8    & 8    & 8    \\
Candidate peaks $N_p$    & 3    & 3    & 3    \\
Window $W$ (s)           & 8    & 20   & 80   \\
\midrule
Forwards/sample $K_c{+}N_pK_f$ & 36 & 40 & 120 \\
Median latency (s)       & 14.4 & 22.4 & 95.6 \\  
\bottomrule
\end{tabular}
\end{table*}

\subsection{Main Results}
\label{sec:main}

Tables~\ref{tab:charades}, \ref{tab:anet}, and \ref{tab:tacos} report results on
Charades-STA, ActivityNet Captions, and TACoS respectively.

\paragraph{Charades-STA.}
FV-Action with Qwen2.5-VL-7B achieves $56.8\%$ R@0.5, surpassing all trained TVG
VLMs including TimeMarker ($51.9\%$), which is trained on over one million TVG samples,
and VideoChat-Flash ($53.1\%$).
Most directly, it surpasses the same backbone's own native temporal pipeline
(Qwen2.5-VL zero-shot, $53.6\%$) on identical frozen weights, and does so on the
tighter R@0.7 as well ($29.7$ vs.\ $28.5$): switching the interface, not the
model, is what improves grounding.
Among training-free methods, FV-Action improves over GranAlign ($39.6\%$) by $17.2$ pp,
despite GranAlign relying on a multi-model pipeline with LLM rewriting and CLIP matching.
It remains competitive with systems that pay for reinforcement learning or large-scale
SFT, trailing Time-R1 ($60.8\%$, RL on 2,500 grounding samples) and
OmniVTG~\cite{zheng2026omnivtg} ($63.2\%$, 359{,}K annotations over 2,124 hours),
while using no temporal training of any kind.
The interface comparison in Table~\ref{tab:diagnostic} shows that every tested
backbone benefits from the VQA interface, with gains from $+28$ to $+50$ pp.

\paragraph{ActivityNet Captions.}
FV-Action achieves $26.7\%$ R@0.5 with Qwen2.5-VL and $25.5\%$ with both
InternVL2-8B and LLaVA-NeXT-Video, surpassing ChatVTG ($22.5\%$), Momentor
($23.0\%$), and UniTime ($22.8\%$), while trailing VTimeLLM ($29.5\%$) and
HawkEye ($29.3\%$), both fine-tuned on temporal grounding data.
The result is stable across the two annotation splits of ActivityNet: the
Qwen2.5-VL R@0.5 is $26.7\%$ on \texttt{val\_1} and $27.0\%$ on \texttt{val\_2}.
Among multi-model training-free methods, GranAlign ($34.0\%$) and BTDP ($30.6\%$)
outperform FV-Action on R@0.5; both use multi-stage pipelines with dense captioners
and LLM alignment, whereas FV-Action uses only a single VLM query.
ActivityNet is also where the adaptive-window variant helps most: it lifts
InternVL2 to $29.8\%$ (Sec.~\ref{sec:ablation}), reflecting the dataset's
heterogeneous event durations.
The primary failure mode under the fixed window is a window/short-GT mismatch:
$38\%$ of queries have GT shorter than 15 s, yet the window is 80 s, causing
structural failures on short events regardless of peak-localization quality.

\paragraph{QVHighlights.}
FV-Action reaches $43.0$ R1@0.5 and $16.7$ R1@0.7 on the val split
($72.9$ R@0.3, mIoU $45.2$; Table~\ref{tab:qvhl}), $1.7\times$ the
zero-shot pretraining transfer of UniVTG ($25.2$) and within eleven points
of the fully supervised Moment-DETR ($53.9$), with no exposure to any
moment-retrieval data. The multi-model pipeline GranAlign is substantially
stronger here ($61.9$), the reverse of the Charades-STA ordering, and the
gap is concentrated at the tight threshold ($41.8$ vs $16.7$ R1@0.7)
where our single fixed $25$\,s width faces moments ranging from seconds
to a minute; QVHighlights is thus the corpus where the geometry axis
costs a single-model scan the most, exactly as the two-axis account
predicts. Perception, in contrast, is our strongest of the four
benchmarks: the selected peak lands inside an annotated moment on
$77.3\%$ of queries.

\begin{table}[t]
\centering
\caption{
  Results on \textbf{QVHighlights} val. Multi-window ground truth is
  scored by maximum IoU (standard protocol). FS: fully supervised on
  QVHighlights; ZS: zero-shot transfer after moment-retrieval pretraining;
  TF: training-free.
}
\label{tab:qvhl}
\setlength{\tabcolsep}{5pt}
\begin{tabular}{l c c c}
\toprule
Method & Setting & R1@0.5 & R1@0.7 \\
\midrule
Moment-DETR~\cite{lei2021detecting} & FS & 53.9 & 34.8 \\
UniVTG~\cite{lin2023univtg}            & ZS & 25.2 &  8.9 \\ 
GranAlign~\cite{jeon2026granalign}     & TF$^\star$ & \textbf{61.9} & \textbf{41.8} \\ 
FV-Action (Qwen2.5-VL-7B)               & TF & 43.0 & 16.7 \\
\bottomrule
\end{tabular}
\end{table}

\paragraph{TACoS.}
FV-Action with Qwen2.5-VL-7B achieves $29.5\%$ R@0.3 and $13.0\%$ R@0.5 on TACoS,
a domain that all compared methods encounter zero-shot.
FV-Action outperforms all trained TVG VLMs on this benchmark, surpassing ChatVTG
($3.7\%$), VTimeLLM ($3.9\%$), ED-VTG ($6.0\%$), and UniTime ($7.4\%$) by $5.7$--$9.4$ pp.
The InternVL2 backbone ($8.6\%$) already exceeds all prior methods; Qwen2.5-VL
provides an additional $+4.6$ pp, reflecting stronger clip-level discriminability.
TACoS remains the hardest setting for FV-Action: long cooking videos (mean 6.7 min)
spread the coarse scan thin, while short GT events compound the fixed-window mismatch.

\begin{table*}[t]
\centering
\caption{
  Results on \textbf{Charades-STA}.
  FS: fully supervised.
  ZS: zero-shot (trained on TVG data from other sources).
  TF: training-free (no TVG supervision at any stage).
  $^\dagger$: uses TVG SFT or pre-training data.
  $^\star$: multi-model pipeline (captioning, LLM rewriting, or retrieval).
  \textbf{Bold}: best among non-FS methods.
  \underline{Underline}: best training-free method.
  The bottom block isolates the interface: each backbone is evaluated under
  direct regression and under FV-Action, with the frozen model held fixed.
  Qwen2.5-VL also appears in the upper block under the native
  temporal-tokenisation interface of~\cite{zheng2026omnivtg} for reference;
  this is an interface-enhanced setting, not a plain generative baseline
  (Sec.~\ref{sec:interface}).
}
\label{tab:charades}
\setlength{\tabcolsep}{4pt}
\begin{tabular}{l c c c c c}
\toprule
Method & Interface & Setting & R@0.3 & R@0.5 & R@0.7 \\
\midrule
\multicolumn{6}{l}{\textit{Fully Supervised}} \\
2D-TAN~\cite{zhang2020learning}      & -- & FS & --   & 45.8 & 27.9 \\
MMN~\cite{wang2022negative}          & -- & FS & --   & 53.3 & 31.5 \\
UniVTG~\cite{lin2023univtg}          & -- & FS & 72.6 & 60.2 & 38.6 \\
\midrule
\multicolumn{6}{l}{\textit{TVG-Trained VLMs (zero-shot on Charades-STA)}} \\
Momentor~\cite{qian2024momentor}     & -- & ZS & 42.6 & 26.6 & 11.6 \\
TimeChat~\cite{ren2024timechat}      & -- & ZS & --   & 31.3 & --   \\
HawkEye~\cite{wang2024hawkeye}       & -- & ZS & 50.6 & 31.4 & 14.5 \\
ChatVTG~\cite{qu2024chatvtg}         & -- & ZS & 52.7 & 33.0 & 15.9 \\
VTG-LLM~\cite{guo2025vtg}            & -- & ZS & --   & 33.8 & 15.7 \\
VTimeLLM~\cite{huang2024vtimellm}    & -- & ZS & 55.3 & 34.3 & 14.7 \\
TRACE~\cite{guo2024trace}            & -- & ZS & --   & 40.3 & 19.4 \\
TimeSuite~\cite{zeng2025timesuite}   & -- & ZS & 69.9 & 48.7 & 24.0 \\
TimeMarker~\cite{chen2024timemarker} & -- & ZS$^\dagger$ & 73.5 & 51.9 & 26.9 \\
VideoChat-Flash~\cite{li2024videochat} & -- & ZS & 74.5 & 53.1 & 27.6 \\
Qwen2.5-VL-7B (native)~\cite{bai2025qwen25vltechnicalreport,zheng2026omnivtg} & -- & ZS$^\dagger$ & 72.5 & 53.6 & 28.5 \\
UniTime~\cite{li2026universal}       & -- & ZS$^\dagger$ & --   & 59.1 & 31.9 \\
Time-R1~\cite{wang2026time}            & -- & ZS$^\dagger$ & 78.1 & 60.8 & 35.3 \\
OmniVTG~\cite{zheng2026omnivtg}      & -- & ZS$^\dagger$ & \textbf{78.3} & \textbf{63.2} & \textbf{37.0} \\
\midrule
\multicolumn{6}{l}{\textit{Training-Free (Multi-Model Pipelines)}} \\
LLM4VG~\cite{feng2023llm4vg}          & -- & TF$^\star$ & --   & 11.8 & --   \\
Moment-GPT~\cite{xu2025zero}  & -- & TF$^\star$ & 58.2 & 38.4 & 21.6 \\
GranAlign~\cite{jeon2026granalign}   & -- & TF$^\star$ & 59.1 & 39.6 & 22.7 \\
BTDP~\cite{deng2025boundary}                 & -- & TF$^\star$ & 58.3 & 40.0 & 20.9 \\
\midrule
\multicolumn{6}{l}{\textit{Training-Free, Single Model (Ours)}} \\
LLaVA-NeXT-Video-7B~\cite{zhang2024llavanextvideo} & Regression & TF & 0.1 &  0.1 &  0.0 \\ 
InternVL2-8B~\cite{chen2024internvl}               & Regression & TF & 32.1 &  3.8 &  1.1 \\
mPLUG-Owl3-7B~\cite{ye2025mplug}             & Regression & TF & 5.6 & 2.3 & 1.0 \\ 
Qwen2.5-VL-7B~\cite{bai2025qwen25vltechnicalreport}               & Regression & TF & 43.0 & 28.4 & 14.6 \\

\cmidrule(l){1-6}
LLaVA-NeXT-Video-7B~\cite{zhang2024llavanextvideo} & VQA-Scan & TF & 59.6 & 47.0 & 23.1 \\
InternVL2-8B~\cite{chen2024internvl}               & VQA-Scan & TF & 66.5 & 53.4 & 26.6 \\
mPLUG-Owl3-7B~\cite{ye2025mplug}             & VQA-Scan & TF & 54.7 & 39.7 & 20.2 \\
Qwen2.5-VL-7B~\cite{bai2025qwen25vltechnicalreport}               & VQA-Scan & TF & \underline{\textbf{70.1}} & \underline{\textbf{56.8}} & \underline{\textbf{29.7}} \\
\bottomrule
\end{tabular}
\end{table*}

\begin{table*}[t]
\centering
\caption{
  Results on \textbf{ActivityNet Captions}.
  Symbols follow Table~\ref{tab:charades}.
  Only the scan interface is evaluated here, so the interface column is
  omitted. Direct regression is reported only on Charades-STA, where the
  frame budgets of the two interfaces match (Sec.~\ref{sec:setup}); on
  ActivityNet the sampling densities differ and a controlled comparison is
  not possible. The geometric account of the gap to TVG-trained models is
  given in Sec.~\ref{sec:analysis}.
}
\label{tab:anet}
\setlength{\tabcolsep}{5pt}
\begin{tabular}{l c c c c}
\toprule
Method & Setting & R@0.3 & R@0.5 & R@0.7 \\
\midrule
\multicolumn{5}{l}{\textit{Fully Supervised}} \\
2D-TAN~\cite{zhang2020learning}      & FS & --   & 43.4 & 25.0 \\
MMN~\cite{wang2022negative}          & FS & --   & 48.2 & 29.4 \\
\midrule
\multicolumn{5}{l}{\textit{TVG-Trained VLMs (zero-shot on ActivityNet)}} \\
ChatVTG~\cite{qu2024chatvtg}         & ZS           & 40.7 & 22.5 &  9.4 \\
UniTime~\cite{li2026universal}       & ZS$^\dagger$ & --   & 22.8 & 14.1 \\
Momentor~\cite{qian2024momentor}     & ZS           & 42.9 & 23.0 & 12.4 \\
HawkEye~\cite{wang2024hawkeye}       & ZS           & 49.1 & 29.3 & 10.7 \\
VTimeLLM~\cite{huang2024vtimellm}    & ZS           & 44.8 & 29.5 & 14.2 \\
Time-R1~\cite{wang2026time}            & ZS$^\dagger$ & 58.6 & 39.0 & 21.4 \\
OmniVTG~\cite{zheng2026omnivtg}      & ZS$^\dagger$ & 60.3 & 39.8 & 21.4 \\
TimeMarker~\cite{chen2024timemarker} & ZS$^\dagger$ & \textbf{67.4} & \textbf{50.7} & \textbf{33.0} \\
\midrule
\multicolumn{5}{l}{\textit{Training-Free (Multi-Model Pipelines)}} \\
Moment-GPT~\cite{xu2025zero}         & TF$^\star$ & 48.1 & 31.1 & 14.9 \\
BTDP~\cite{deng2025boundary}         & TF$^\star$ & \underline{50.6} & 30.6 & \underline{17.5} \\
GranAlign~\cite{jeon2026granalign}   & TF$^\star$ & 50.3 & \underline{34.0} & 16.5 \\
\midrule
\multicolumn{5}{l}{\textit{Training-Free, Single Model (Ours)}} \\
FV-Action (mPLUG-Owl3-7B)            & TF & 41.0 & 21.6 &  9.6 \\ 
FV-Action (LLaVA-NeXT-Video-7B)      & TF & 45.9 & 25.5 & 11.2 \\ 
FV-Action (InternVL2-8B)             & TF & 46.7 & 25.5 & 11.2 \\ 
FV-Action (Qwen2.5-VL-7B)            & TF & 48.6 & 26.7 & 11.9 \\ 
\bottomrule
\end{tabular}
\end{table*}

\begin{table*}[t]
\centering
\caption{
  Results on \textbf{TACoS}.
  All non-FS methods are evaluated zero-shot with respect to TACoS
  training data. Symbols follow Table~\ref{tab:charades}; the interface
  column is omitted as only the scan interface is evaluated.
}
\label{tab:tacos}
\setlength{\tabcolsep}{5pt}
\begin{tabular}{l c c c c}
\toprule
Method & Setting & R@0.3 & R@0.5 & R@0.7 \\
\midrule
\multicolumn{5}{l}{\textit{Fully Supervised}} \\
2D-TAN~\cite{zhang2020learning}      & FS & 37.3 & 25.3 & --  \\
MMN~\cite{wang2022negative}          & FS & 39.2 & 26.6 & --  \\
UniVTG~\cite{lin2023univtg}          & FS & 51.4 & 35.0 & --  \\
\midrule
\multicolumn{5}{l}{\textit{TVG-Trained VLMs (zero-shot on TACoS)}} \\
TimeChat~\cite{ren2024timechat}      & ZS           & --   &  2.1 & --  \\
ChatVTG~\cite{qu2024chatvtg}         & ZS           &  8.1 &  3.7 & 1.3 \\
VTimeLLM~\cite{huang2024vtimellm}    & ZS           & --   &  3.9 & --  \\
ED-VTG~\cite{pramanick2025enrich}    & ZS           & 14.5 &  6.0 & 2.3 \\
UniTime~\cite{li2026universal}       & ZS$^\dagger$ & --   &  7.4 & --  \\
\midrule
\multicolumn{5}{l}{\textit{Training-Free, Single Model (Ours)}} \\
FV-Action (LLaVA-NeXT-Video-7B)      & TF & 12.1 &  6.4 & 1.2 \\ 
FV-Action (mPLUG-Owl3-7B)            & TF & 19.4 &  7.6 & 1.2 \\ 
FV-Action (InternVL2-8B)             & TF & 20.8 &  8.6 & 1.9 \\ 
FV-Action (Qwen2.5-VL-7B)            & TF & \textbf{\underline{29.5}} & \textbf{\underline{13.0}} & \textbf{\underline{3.3}} \\
\bottomrule
\end{tabular}
\end{table*}

\subsection{The Interface Bottleneck}
\label{sec:interface}

\paragraph{Same model, different interface.}
Table~\ref{tab:diagnostic} isolates the effect of task formulation.
Both interfaces are training-free and use identical input, sixteen uniformly
sampled frames with a simple zero-shot prompt, no model-specific temporal
encoding and no TVG supervision; the only variable is whether the model emits
timestamps or answers a binary VQA query.
Across every backbone the VQA interface outperforms timestamp regression by
$28$--$50$ pp in R@0.5.
InternVL2-8B rises from $3.76$ under regression to $53.39$ with scanning: under
the generative interface it degenerates to emitting the whole video ($[0,T]$),
which reaches R@0.3 by trivial coverage but almost never R@0.5, even though its
clip-level discriminability reaches AUROC $0.782$ on the same frames, so the
missing ingredient is not perception but the output channel.
LLaVA-NeXT-Video-7B is the extreme case: under regression it fails to emit two
parseable timestamps on $69\%$ of samples, so scored strictly its R@0.5 is
$0.05$, yet the same frozen backbone recovers to $47.04$ with scanning.
Qwen2.5-VL-7B rises from $28.41$ to $56.77$.
Because the backbone is frozen and only the output format changes, this gap
identifies the interface as the cause.
The comparison is sharpest against the native dense temporal pipeline of
Qwen2.5-VL ($53.6$ R@0.5~\cite{zheng2026omnivtg}): on the same weights,
FV-Action exceeds it on both R@0.5 ($56.8$) and R@0.7 ($29.7$ vs.\ $28.5$)
using only sixteen frames and no temporal machinery.

\begin{table*}[t]
\centering
\caption{
  \textbf{The interface bottleneck: same backbone, same frames, different
  interface.}
  Both rows per backbone are training-free and use identical input
  (16 uniformly sampled frames, zero-shot prompting, no TVG supervision).
  The only difference is whether the model outputs timestamps or answers
  binary VQA queries. Charades-STA R@$\tau$.
}
\label{tab:diagnostic}
\setlength{\tabcolsep}{4pt}
\begin{tabular}{l l c c c c}
\toprule
Backbone & Interface & R@0.3 & R@0.5 & R@0.7 & $\Delta$R@0.5 \\
\midrule
\multirow{2}{*}{InternVL2-8B}
  & Direct (timestamp)    & 32.12 &  3.76 &  1.08 & \multirow{2}{*}{$+49.63$ pp} \\
  & FV-Action VQA (ours)  & \textbf{66.51} & \textbf{53.39} & \textbf{26.64} & \\
\midrule
\multirow{2}{*}{Qwen2.5-VL-7B}
  & Direct (timestamp)    & 43.04 & 28.41 & 14.57 & \multirow{2}{*}{$+28.36$ pp} \\
  & FV-Action VQA (ours)  & \textbf{70.13} & \textbf{56.77} & \textbf{29.68} & \\
\midrule
\multirow{2}{*}{LLaVA-NeXT-Video-7B}
  & Direct (timestamp)    &  0.08 &  0.05 &  0.00 & \multirow{2}{*}{$+46.99$ pp} \\
  & FV-Action VQA (ours)  & \textbf{59.62} & \textbf{47.04} & \textbf{23.06} & \\
\midrule
\multirow{2}{*}{mPLUG-Owl3-7B}
  & Direct (timestamp)    & 5.62 & 2.31 & 	1.02 & \multirow{2}{*}{$+37.37$ pp} \\
  & FV-Action VQA (ours)  & \textbf{54.70} & \textbf{39.68} & \textbf{20.22} & \\
\bottomrule
\end{tabular}
\end{table*}

\paragraph{Is the regression baseline under-prompted?}
A gap this large invites the objection that the direct baseline simply
received a poor prompt. Table~\ref{tab:promptvar} answers it by sweeping five
prompt formulations for Qwen2.5-VL under the regression interface, holding
frames, decoding and parser fixed, on a common $400$-sample subset. The prompt
deployed in Table~\ref{tab:diagnostic}, which states the video duration, is the
best of the five and is reproduced by variant~V1 to within half a point of
R@0.5. Withholding the duration (V0) costs $9.8$ pp, and the two formulations
that supply the model with \emph{more} temporal scaffolding are the two that
hurt most: per-frame timestamps in the text (V2) drop format compliance to
$66.8\%$, and chain-of-thought reasoning before the answer (V3) destroys it
entirely, with no sample in the subset emitting a parseable interval. The
deployed baseline therefore sits at the maximum of this family rather than
below it, and the residual gap to VQA scanning is not a prompting artefact.
The V3 result is the sharpest form of the interface problem: asking the model
to reason about time is precisely what stops it from reporting time in a
format anyone can read.

\begin{table}[t]
\centering
\caption{
  \textbf{Prompt sensitivity of the regression baseline} (Qwen2.5-VL-7B,
  Charades-STA, common $400$-sample subset, 16 frames, identical parser).
  \emph{Fmt} is the fraction of outputs matching the requested
  \texttt{start/end} format; unparseable outputs score IoU${=}0$.
  V1 is the formulation deployed in Table~\ref{tab:diagnostic}.
}
\label{tab:promptvar}
\setlength{\tabcolsep}{4pt}
\footnotesize
\begin{tabular}{l l c c c c}
\toprule
& Prompt & R@0.3 & R@0.5 & R@0.7 & Fmt \\
\midrule
V0 & no duration given        & 40.75 & 27.25 & 12.75 & $80.2\%$ \\
V1 & \textbf{+ video duration (deployed)} & \textbf{55.75} & \textbf{37.00} & \textbf{19.75} & $\mathbf{95.5\%}$ \\
V2 & + per-frame timestamps   & 30.25 & 18.50 & 10.75 & $66.8\%$ \\
V3 & chain-of-thought first   &  0.00 &  0.00 &  0.00 & $0.0\%$ \\
V4 & integer seconds          & 41.25 & 26.50 & 12.50 & $80.8\%$ \\
\bottomrule
\end{tabular}
\end{table}

\paragraph{Is the scanning interface prompt-sensitive too?}
The converse question is whether the VQA arm merely hides its own prompt
sensitivity. Table~\ref{tab:vqaprompt} answers it by rewriting the scan
question five ways under a frozen pipeline, on the same $400$-sample subset.
The spread is $3.2$ pp R@0.5 between the best and worst formulation, against
$18.5$ pp on the regression side of Table~\ref{tab:promptvar}.
We state what this subset can and cannot establish. At $n{=}400$ the paired
McNemar test resolves differences of about $5$ pp, so the result bounds the
effect rather than excluding it: no rewriting we tried moves R@0.5 by the
double-digit margin that separates the two interfaces, but a small effect
would not be visible here. The largest gain on the subset, dropping the
\texttt{Answer Yes or No} suffix (S5), was $+2.0$ pp R@0.5 but with a $95\%$
confidence interval of $[-1.5, +5.5]$; re-running S5 against a paired baseline
on the full $3{,}720$ samples, where the test resolves $1.7$ pp, collapses it
to $-0.4$ pp R@0.5 ($p{=}0.57$) and $0.0$ pp R@0.7 ($p{=}1.0$). The apparent
subset gain was sampling noise, and we keep the deployed prompt.
One hypothesis is settled rather than merely unresolved. Charades queries
begin with a bare \texttt{person} on $57.5\%$ of samples, so the deployed
template produces ungrammatical questions (``\emph{Does this clip show person
turn a light on?}''). Repairing the article perturbs the scan curve on every
one of those samples, yet the argmax survives on $81.3\%$ of them and R@0.5
and peak-in-GT are numerically identical, with only $8$ of $400$ samples
crossing the R@0.5 threshold in either direction (confidence interval
$[-1.5, +1.5]$). Here the tight interval does license the negative reading:
grammar shifts the scores without reordering them, and the scan consumes only
the order.
We also let the backbone rewrite its own question rather than relying on our
hand-written variants (S6): Qwen2.5-VL is prompted to turn each query into a
clean binary question, which it does with no format failures, and the pipeline
runs on the result. This does not help either, and mildly hurts R@0.7
($-4.3$ pp, $p{=}0.06$, against a paired baseline of $32.8$), because the
rewrites tend to over-specify a loosely annotated action; even an automated
search over phrasings finds no headroom.
That mechanism is the interface claim restated. Ranking absorbs perturbations
that would corrupt a generated string, which is why phrasing costs the scan at
most a few points while the same backbone asked to emit timestamps swings by
$18.5$ pp on wording alone. We note that the two sweeps are not perturbations
of equal severity: the regression variants include structural changes such as
interleaved per-frame timestamps and chain-of-thought, whereas the scan
variants are rewordings of a fixed question, since the binary format admits no
structural analogue.

\begin{table}[t]
\centering
\caption{
  \textbf{Prompt sensitivity of the scanning interface} (Qwen2.5-VL-7B,
  Charades-STA, $400$-sample subset; pipeline frozen, only the scan question
  varies). \emph{Peak} is the fraction of located peaks falling inside the
  ground-truth interval. At this subset size a paired test resolves about
  $5$ pp, so the table bounds prompt sensitivity rather than excluding it;
  contrast the $18.5$ pp spread in Table~\ref{tab:promptvar}.
}
\label{tab:vqaprompt}
\setlength{\tabcolsep}{4pt}
\footnotesize
\begin{tabular}{l l c c c c}
\toprule
& Scan question & R@0.3 & R@0.5 & R@0.7 & Peak \\
\midrule
S0 & \textbf{deployed}            & 71.8 & 58.0 & 32.8 & $69.0\%$ \\
S1 & + article repair            & 71.3 & 58.0 & 31.8 & $69.0\%$ \\
S2 & present progressive         & 70.5 & 57.3 & 32.3 & $68.8\%$ \\
S3 & ``happening at this moment'' & 69.8 & 57.3 & 33.5 & $67.0\%$ \\
S4 & subject removed             & 69.0 & 56.8 & 32.0 & $67.3\%$ \\
S5 & no \texttt{Yes/No} suffix   & 72.8 & 60.0 & 32.0 & $70.3\%$ \\
S6 & model-rewritten (Qwen)      & 70.0 & 57.3 & 28.5 & $68.8\%$ \\
\bottomrule
\end{tabular}
\end{table}

\paragraph{Backbone generality.}
The interface gain is not a property of one model. Table~\ref{tab:diagnostic}
shows all four backbones, spanning four architectures, improving by $28$ to
$50$ pp R@0.5 over their own regression baselines under the identical frozen
pipeline, with the backbone ordering preserved across the switch.

\section{Analysis: Two Axes of Failure}
\label{sec:analysis}

\begin{figure*}[t]
\centering
\includegraphics[width=\textwidth]{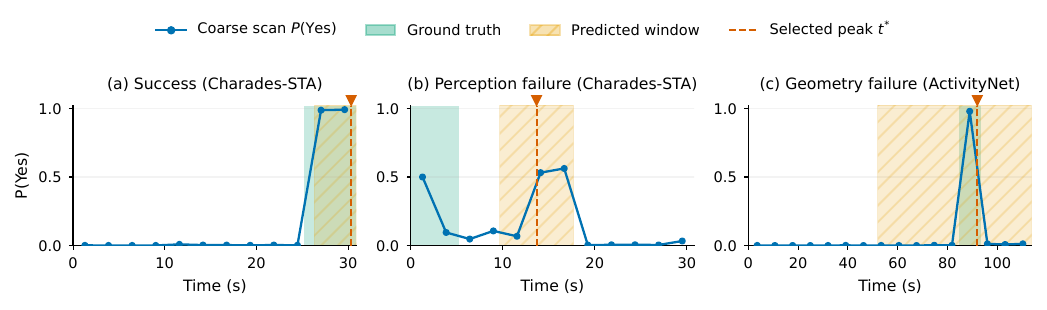}
\caption{
  \textbf{The scan in practice, one example per regime} (Qwen2.5-VL coarse
  curves; ground truth shaded, predicted window hatched, selected peak
  dashed).
  (a) \emph{``person sits on a couch''}: a sharp peak inside the event and
  a well-placed $W{=}8$\,s window (IoU $0.80$).
  (b) \emph{``one person opens the door''}: the true event (left band)
  scores $0.50$ but a visually similar later moment scores $0.56$; the peak
  lands there and the window misses. The margin of $0.06$ is typical of the
  small-margin perception noise characterised in
  Sec.~\ref{sec:perception}.
  (c) \emph{``powdered sugar is sprinkled over the cake''}: the peak lands
  inside the $8.5$\,s event, yet the fixed $W{=}80$\,s window dilutes IoU
  to $0.14$. Perception succeeded and geometry failed, the separation the
  two-axis analysis quantifies.
}
\label{fig:qualitative}
\end{figure*}

Where FV-Action still fails, the failures decompose along two measurable and
separable axes: a \emph{perception} axis governed by the backbone's per-clip
discriminability, and a \emph{geometry} axis governed by the fixed output
window. Figure~\ref{fig:qualitative} shows one example of each regime.
We characterise each axis, show they are separable, turn the
perception axis into a pre-deployment diagnostic, catalogue the alternatives
that do not work, and close with component ablations.

\subsection{The Perception Axis: Backbone-Dependent Discriminability}
\label{sec:perception}

\paragraph{Oracle protocol.}
For each sample we take the ground-truth interval as a positive clip and an
equal-width non-overlapping window from the same video as a negative, and
record the signal gap
$\Delta = P(\text{Yes}\mid\text{GT}) - P(\text{Yes}\mid\text{neg})$ and the
discrimination rate $\mathrm{disc} = P(p_\text{gt} > p_\text{neg})$.
This measures per-clip separability, what the scan actually consumes,
independent of any localisation machinery.

\paragraph{The axis moves with the backbone.}
On TACoS the oracle gap rises from $0.186$ (InternVL2) to $0.260$
(Qwen2.5-VL), a paired difference on identical samples
($n{=}200$, $p{=}0.0009$), and pipeline R@0.5 rises with it from $8.6$ to
$13.0$ with nothing else changed.
Entropy offers no comparable handle: it discriminates correct from incorrect
predictions at AUROC $0.539$--$0.578$ across TimeChat, LITA, and VTG-LLM, near
chance, whereas the clip-level VQA probability reaches AUROC $0.782$ on
InternVL2-8B, which is why entropy-gated correction reaches only sub-random
F1 (below $0.36$) while VQA scanning succeeds.

\paragraph{Ranking, not calibration.}
On TACoS, Qwen2.5-VL answers the oracle question correctly on only $32.5\%$ of
ground-truth clips ($\bar p_\text{gt} = 0.372$), yet its discrimination rate
is $80.5\%$ because negatives sit on a clean low baseline
($\bar p_\text{neg} = 0.112$).
A backbone can therefore appear poorly calibrated in isolation and still be a
strong scanner, precisely the property the pipeline consumes
(Sec.~\ref{sec:stage1}).

\paragraph{The residual sits on the perception axis, not the pipeline.}
A natural question is whether smarter curve processing could recover the
located peak where it currently misses. On Charades-STA the peak lands inside
the ground truth on $67.7\%$ of samples, and among the failures $65.6\%$ have
the best in-interval clip ranked within the coarse top three, which would
suggest a ceiling of $88.8\%$ peak accuracy if a selection rule could exploit
it. That headroom is illusory. The wrong peak beats the best in-interval clip
by a median of only $0.093$ in $P(\text{Yes})$: the errors are small-margin
perception noise, not a systematic bias a filter can subtract. Accordingly,
every curve-level remedy we tried lowers accuracy rather than raising it.
Temporal smoothing of the coarse curve drops peak accuracy to $62.4\%$ and
R@0.5 to $52.6$; contrastive scoring, which subtracts a background-query
probability to suppress scene bias, drops them to $62.4\%$ and $54.1$;
doubling the probe budget and rewording the question leave both unchanged
(Sec.~\ref{sec:ablation}, Table~\ref{tab:vqaprompt}). What does move peak
accuracy is the backbone: on the identical pipeline it rises monotonically
$50.9 \rightarrow 57.0 \rightarrow 63.1 \rightarrow 67.7\%$ across mPLUG-Owl3,
LLaVA-NeXT-Video, InternVL2, and Qwen2.5-VL, tracking their R@0.5 order. The
two-axis account is therefore not only descriptive but prescriptive: with
geometry handled by the window prior, the remaining error is a property of
per-clip discriminability, and the operative lever is a stronger instrument,
not a cleverer read of a fixed one.

\subsection{The Geometry Axis: Analytically Predictable Coverage}
\label{sec:geometry}

\begin{figure}[t]
  \centering
  \includegraphics[width=\linewidth]{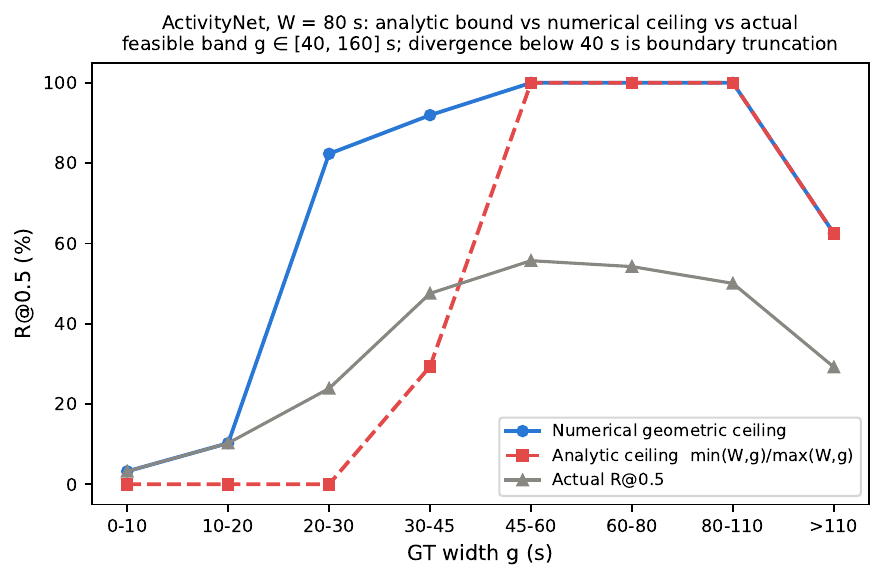}
  \caption{
    \textbf{The analytic bound predicts failure before any model runs.}
    ActivityNet, $W{=}80$\,s, by ground-truth width $g$.
    The closed form $\min(W,g)/\max(W,g) \geq 0.5$ admits only
    $g \in [40, 160]$\,s.
    Buckets below the band collapse as predicted; the residual non-zero
    values there are boundary truncation, where the window is clipped at
    the video edge and its effective width shrinks.
  }
  \label{fig:analytic}
\end{figure}

\begin{figure}[t]
  \centering
  \includegraphics[width=\linewidth]{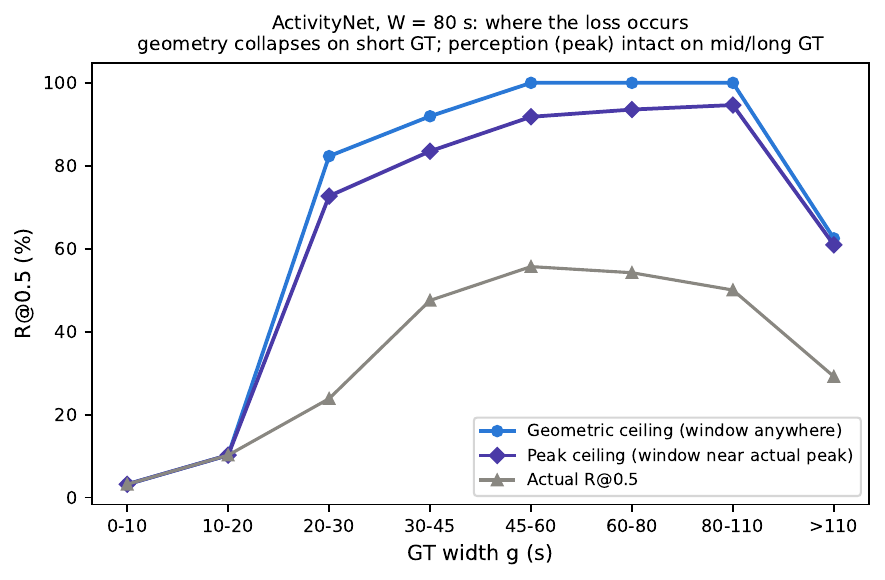}
  \caption{
    \textbf{Where the loss occurs.}
    ActivityNet, $W{=}80$\,s.
    On mid and long buckets the peak ceiling stays at $61$--$95\%$ while
    actual R@0.5 is $29$--$56\%$: the peak is found, the window cannot
    express it. On short buckets all three curves coincide: the failure is
    purely geometric.
  }
  \label{fig:threeline}
\end{figure}

\begin{figure}[t]
  \centering
  \includegraphics[width=\linewidth]{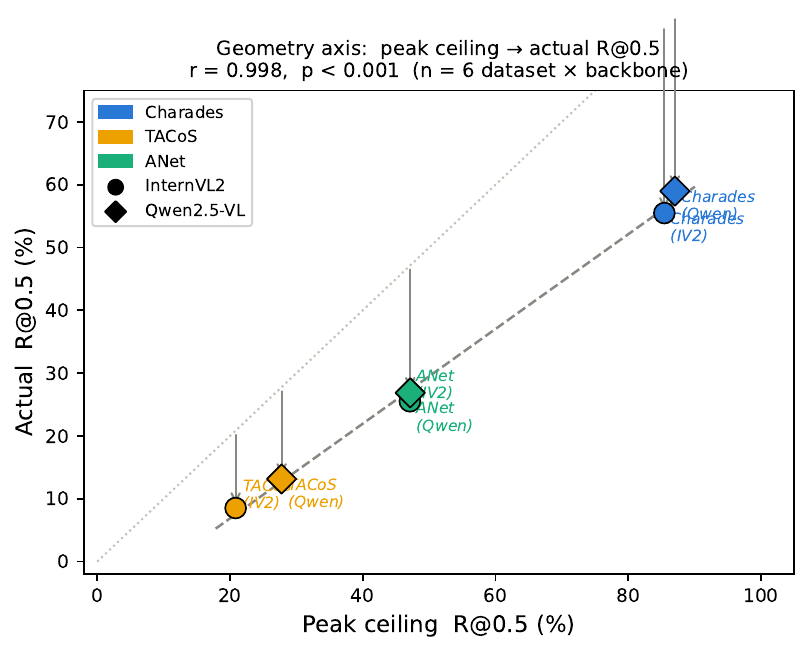}
  \caption{
    \textbf{Peak ceiling explains final accuracy.}
    Across six dataset and backbone combinations the geometric ceiling
    computed from actual peak positions tracks final R@0.5 with
    $r = 0.998$; the vertical drop to the diagonal is the windowing loss.
  }
  \label{fig:scatter}
\end{figure}

Eq.~\eqref{eq:geo} implies a feasible band $g \in [W/2,\, 2W]$ outside of
which R@0.5 is zero for a perfectly placed window.
Figure~\ref{fig:analytic} tests this on ActivityNet at $W{=}80$\,s: buckets
with ground truth shorter than $40$\,s collapse to $3$--$24\%$ exactly as the
bound dictates, with the small residuals explained by boundary truncation,
while buckets inside the band reach $50$--$56\%$.
Figure~\ref{fig:threeline} separates the two axes on the same data: for $g$
between $45$ and $110$\,s the peak ceiling holds at $92$--$95\%$, so
perception has already succeeded and the gap to the actual $50$--$56\%$ is
entirely the window's inability to express the found peak.
Aggregating over all six dataset and backbone combinations, the peak ceiling
predicts final R@0.5 with $r = 0.998$ (Fig.~\ref{fig:scatter}); the residual
to the diagonal is the geometry loss.

The adaptive-width variant is the direct response to this axis.
Table~\ref{tab:awp} shows its per-bucket effect on ActivityNet with
Qwen2.5-VL: adaptation gains $10$--$18$ pp on both tails ($g<30$\,s and
$g>110$\,s) where the fixed width mismatches the event, and loses $12$--$21$
pp on the matched middle, for a net $+3.4$ pp; the pattern is stable under
both an absolute and a baseline-corrected half-maximum definition.
A remedy for the geometry axis thus exists but carries a two-sided
trade-off (Sec.~\ref{sec:negative}).

\begin{table}[t]
\centering
\caption{
  \textbf{Two-sided effect of the adaptive width} (ActivityNet, Qwen2.5-VL,
  R@0.5 by ground-truth width). Adaptation helps where fixed $W{=}80$\,s
  mismatches the event width and hurts where it matches.
}
\label{tab:awp}
\setlength{\tabcolsep}{4pt}
\begin{tabular}{l c c c c}
\toprule
GT width & $n$ & Fixed $W$ & Adaptive & $\Delta$ \\
\midrule
0--10\,s    & 4305 &  3.3 & 16.2 & $+12.9$ \\
10--20\,s   & 2795 & 10.5 & 28.3 & $+17.7$ \\
20--30\,s   & 1901 & 25.3 & 34.9 & $+9.6$ \\
30--45\,s   & 1914 & 52.4 & 38.0 & $-14.3$ \\
45--60\,s   & 1259 & 59.0 & 38.0 & $-21.0$ \\
60--80\,s   & 1102 & 58.4 & 43.2 & $-15.2$ \\
80--110\,s  &  939 & 51.5 & 39.5 & $-12.0$ \\
$>$110\,s   & 1004 & 29.6 & 39.4 & $+9.9$ \\
\bottomrule
\end{tabular}
\end{table}

\subsection{Axis Separability}
\label{sec:separability}

Three observations establish that the two axes are independent.
First, geometry can fail while perception is intact: on the mid and long
ActivityNet buckets the peak ceiling holds at $92$--$95\%$ while actual
accuracy is $50$--$56\%$ (Fig.~\ref{fig:threeline}).
Second, when perception is equivalent, geometry decides. An earlier snapshot
of our results appeared to show InternVL2 ahead of Qwen2.5-VL on ActivityNet
($29.8$ vs.\ $24.5$), but the two runs used different window policies (an
adaptive width against a fixed $W{=}38$\,s); under the same fixed $W{=}80$\,s
the ordering reverses ($25.5$ vs.\ $26.7$), and under the same adaptive
variant the two fall within one point ($29.2$ vs.\ $30.2$). The apparent
backbone gap was a window-policy artefact.
Third, when perception fails, geometry cannot compensate. On TACoS the
analytic probability that some probe lands in the ground truth is $0.91$ at
$K_c{=}96$, yet the measured peak hit rate is only $27$--$39\%$: the probes
reach the event but per-clip scores do not lift it above the noise peaks, and
no window rule can recover a peak that was never selected.

\subsection{Predicting Deployment: The Oracle-Gap Diagnostic}
\label{sec:diagnostic}

The oracle gap of Sec.~\ref{sec:perception} is measured from two hundred
clips with no localisation, making it a cheap pre-deployment probe of the
perception axis. Where that axis binds, the gap predicts which backbone will
deploy better: on TACoS the gap ordering (InternVL2 $0.186 <$ Qwen2.5-VL
$0.260$) matches the deployed ordering ($8.6 < 13.0$) before the full
pipeline is ever run.
Two scope conditions keep the probe honest. First, it predicts only in the
perception-limited regime; where geometry binds instead, as on Charades-STA
where both backbones already discriminate above $73\%$, the gap does not
determine R@0.5 and the peak ceiling of Sec.~\ref{sec:geometry} does.
Second, the gap does not transfer across datasets: it inflates with the
clip-to-event width ratio, and re-measuring it at each pipeline's deployment
clip width raises the cross-dataset correlation with the peak ceiling only to
$r{=}0.47$ ($p{=}0.43$, $n{=}5$).
The gap is therefore a within-dataset diagnostic for the perception axis, and
its failure to transfer is itself evidence for the two-axis decomposition:
no single scalar predicts deployment because two independent axes govern it.

\subsection{What Does Not Work}
\label{sec:negative}

Table~\ref{tab:negative} compiles the alternatives we tested and rejected;
each eliminated a hypothesis and shaped one design choice.
The most consequential is the curve-derived gating rule that would apply the
adaptive width of Sec.~\ref{sec:geometry} only where it helps: because a
clip's scan-curve width (its full width at half maximum) is uncorrelated with
the event duration $g$, no global threshold recovers the adaptive gain
without reintroducing the fixed-window loss, and the three datasets demand
opposite thresholds. The event width is simply not encoded in the per-clip
signal, which is why the geometry axis cannot be closed from the curve and the
fixed window remains the default.
A second result explains why the coarse scan places its probes uniformly
rather than at visually salient instants. Replacing the uniform clip centres
with content-adaptive key frames, chosen as the largest inter-frame
differences so that no extra model is introduced, costs $13.1$ pp R@0.5 on
Charades-STA ($43.7$ vs.\ $56.8$) and drops the fraction of peaks landing
inside the ground-truth interval from $67.7\%$ to $53.7\%$, on identical
samples and an otherwise identical pipeline. Large visual change marks camera
motion and lighting shifts as readily as action onsets, so key-frame selection
spends the probe budget on the wrong instants. Given a fixed budget and no
reliable prior over where the event lies, spreading probes uniformly is the
better strategy, and the fine scan then buys locality where a coarse peak has
already been found.

\begin{table}[t]
\centering
\caption{
  \textbf{Negative results.} Each rejected alternative is reported with the
  hypothesis it eliminated.
}
\label{tab:negative}
\setlength{\tabcolsep}{3pt}
\footnotesize
\begin{tabular}{p{0.30\linewidth} p{0.28\linewidth} p{0.32\linewidth}}
\toprule
Alternative & Outcome & What it rules out \\
\midrule
Curve-derived gating of the adaptive window &
no global threshold passes; datasets demand opposite $\theta$ &
Event width is not encoded in the scan curve (FWHM $\perp g$); the geometry
axis cannot be closed from the curve \\
\addlinespace[2pt]
Interval estimation from thresholded score support &
$15.9$ vs.\ $56.8$ R@0.5 on Charades; worse on all three corpora &
Boundary estimation needs signal quality the per-clip scores lack; justifies
peak-plus-window output \\
\addlinespace[2pt]
Per-video adaptive probe budget (denser on short videos) &
no gain over fixed $K_c$ on ActivityNet &
Redistributing probes cannot help once uniform doubling already does not
(full-set result above); the limit is perception, not sampling \\
\addlinespace[2pt]
Content-adaptive (key-frame) probe placement &
$43.7$ vs.\ $56.8$ R@0.5; peak-in-GT $53.7\%$ vs.\ $67.7\%$ &
Visual change is not action onset; uniform placement is the right prior for
where to spend a fixed probe budget \\
\addlinespace[2pt]
Boundary and onset queries (``does it start here'') &
$-45$ pp vs.\ main &
VLMs cannot reliably answer inception-style questions \\
\addlinespace[2pt]
Query decomposition into sub-events &
$-18$ pp &
Sub-event peaks do not separate on the scan curve; multi-event queries remain
a limitation \\
\bottomrule
\end{tabular}
\end{table}

\subsection{Ablations}
\label{sec:ablation}

Table~\ref{tab:ablation} changes one component at a time under the main
configuration.
Removing the fine scan (localising with the coarse argmax alone) costs $1.2$
pp on Charades but is neutral to slightly positive on TACoS and ActivityNet,
where the coarse grid is already dense relative to the event scale; the fine
scan earns its place on short-video corpora.
Replacing the fixed window with the adaptive FWHM variant reproduces the
two-sided behaviour of Sec.~\ref{sec:geometry}: a large loss on Charades and
gains on TACoS and ActivityNet.
Swapping the backbone from Qwen2.5-VL to InternVL2-8B moves every dataset,
most strongly TACoS, which Sec.~\ref{sec:perception} traces to per-clip
discriminability.

\begin{table}[t]
\centering
\caption{
  \textbf{Component ablations} (R@0.5). Each row changes one element of the
  main configuration (Qwen2.5-VL, fixed $W$, Top-$N_p$ NMS with joint
  scoring).
}
\label{tab:ablation}
\setlength{\tabcolsep}{4pt}
\begin{tabular}{l c c c}
\toprule
Variant & Charades & TACoS & ANet \\
\midrule
FV-Action (full)                    & \textbf{56.77} & 13.00 & 26.69 \\
$-$ fine scan (coarse argmax)       & 55.54 & 13.95 & 26.46 \\
fixed $W \rightarrow$ FWHM adaptive & 51.72 & \textbf{18.75} & \textbf{30.24} \\
backbone $\rightarrow$ InternVL2-8B & 53.39 &  8.52 & 25.52 \\
\bottomrule
\end{tabular}
\end{table}

The remaining settings divide into two families with opposite verdicts.
Two input-evidence choices, the frames each probe reads and the resolution
it reads them at, genuinely move accuracy, and we report them first.
Every algorithmic setting after that sits on a measured plateau, which is
the quantitative form of the claim that nothing in the pipeline is tuned
to the test set.

\paragraph{Frames per probe.}
Every configuration above reads $F{=}3$ frames per probe, a default inherited
from the first prototype and, like the prompt wording, never revisited.
Sweeping it exposes the one Stage-1 setting that does move localisation
quality at a fixed backbone. Raising $F$ to five frames while widening the
probe's temporal extent from $2\delta$ to $4\delta$, where $\delta$ is one
third of the probe spacing, and extracting frames symmetrically about the
probe centre improves every metric on the full Charades-STA test set: R@0.5
rises from $56.7$ to $62.3$ ($+5.7$ pp, paired $p{=}1.4\times10^{-12}$,
$n{=}3720$), R@0.7 from $29.5$ to $31.9$ ($p{=}1.1\times10^{-3}$), and the
peak-hit rate from $67.4\%$ to $72.7\%$ ($p{=}8\times10^{-13}$). The two
ingredients contribute separately. Five frames at the unchanged extent gain
$+2.6$ pp R@0.5 ($p{=}1.2\times10^{-3}$), and widening the extent adds a
further $+3.1$ pp ($p{=}6\times10^{-5}$). Unlike every transformation of the
score curve tested in Section~\ref{sec:perception}, this lever changes the
evidence each probe sees rather than how the curve is post-processed, which
is exactly where the peak-hit diagnosis says the remaining headroom lies.

The gain has a sweet spot on both sides. Eight frames over the same
$4\delta$ extent change nothing (largest $|\Delta|$ of $1.0$ pp across all
thresholds, none significant, full test set), so evidence saturates once
within-probe frame spacing reaches the sub-second scale. Eight frames spread
over $7\delta$ hurt ($-2.6$ pp R@0.5, $-4.1$ pp R@0.7, both $p{\le}8\times
10^{-4}$), as the probe's span grows toward the output-window width and
off-event frames dilute the judgement.

The effect generalises with a magnitude that tracks how much of an event a
probe can cover (Table~\ref{tab:fsweep}). At $F{=}5$ the within-probe frame
spacing is $0.85$\,s on Charades, $1.1$\,s on TACoS, and $2.6$\,s on
ActivityNet (median-length videos, deployed $K_c$). Charades events, with a
mean length of $7.8$\,s, are the best matched to the enriched probes and gain
the most; TACoS gains substantially on R@0.3 and R@0.5; ActivityNet, whose
minute-scale events already exceed any probe span and whose residual sits on
the geometry axis, gains least, yet the improvement remains significant at
$n{=}17{,}373$ and no dataset regresses on any threshold.

\begin{table*}[t]
\centering
\caption{
  \textbf{Frames per probe} (Qwen2.5-VL, paired on identical samples, all
  other settings fixed). $F{=}5$ with a $4\delta$ extent versus the $F{=}3$
  default. Positive deltas favour $F{=}5$; $^{*}$ marks paired McNemar
  $p{<}0.01$. No dataset regresses on any threshold.
}
\label{tab:fsweep}
\setlength{\tabcolsep}{4pt}
\begin{tabular}{l c c c c c}
\toprule
Dataset & $n$ & $\Delta$R@0.3 & $\Delta$R@0.5 & $\Delta$R@0.7 & $\Delta$Peak-hit \\
\midrule
Charades-STA & 3720 & $+5.1^{*}$ & $+5.7^{*}$ & $+2.4^{*}$ & $+5.3^{*}$ \\
TACoS        & 3951 & $+6.1^{*}$ & $+3.0^{*}$ & $+0.3$ & $+6.6^{*}$ \\
ActivityNet  & 17373 & $+1.0^{*}$ & $+0.8^{*}$ & $+0.3$ & $+1.2^{*}$ \\
\bottomrule
\end{tabular}
\end{table*}

\paragraph{Input resolution.}
The extra frames also change where the compute budget is best spent. All
runs so far use $448^2$ inputs, and since vision tokens scale with pixel
count, resolution is the largest cost knob after $K_c$. At $F{=}3$ it cannot
be turned down safely: dropping to $224^2$ costs $6.0$ pp R@0.5 (paired
$p{=}0.037$, $n{=}400$). At $F{=}5$ the additional frames absorb most, but
not all, of the loss. On the full test set ($n{=}3720$) the drop to $224^2$
still costs about two points on every threshold ($-2.4$ pp R@0.5, $-1.6$ pp
R@0.7, all $p{<}0.02$), while $336^2$ stays within $1.3$ pp of $448^2$ with
no significant difference on any threshold (closest $p{=}0.053$ at R@0.7).
A $400$-sample pilot had suggested $224^2$ was free; the full set overturns
that reading, which is why we treat subset results as bounds rather than
verdicts. We keep $448^2$ as the deployed resolution. The speed option
remains real: $224^2$ at $F{=}5$ runs at $62\%$ of the $448^2$, $F{=}3$
latency yet still exceeds that deployment by $3.3$ pp R@0.5, so when
latency dominates, resolution is the right knob to turn precisely because
the five-frame probes absorb most of the cost.

The remaining paragraphs sweep every other setting in the pipeline.
The through-line is uniform: each sits on a plateau, the deployed value
lies on it, and where a nominally better point exists it is statistically
indistinguishable, so we keep the prior-derived or default choice.

\paragraph{Joint-score weighting.}
The joint score combines the coarse and fine evidence as a geometric mean
with weight $\alpha{=}0.5$. Recomputing peak selection offline over the
stored candidate curves for $\alpha \in \{0, 0.1, \ldots, 1\}$ on the full
Charades-STA test set traces a single smooth plateau rather than a knife
edge. R@0.5 is flat within $0.3$ pp for $\alpha \in [0.3, 0.6]$, the nominal
maximum at $\alpha{=}0.4$ ($57.02$) is not distinguishable from the deployed
$\alpha{=}0.5$ ($56.75$; paired $p {\ge} 0.14$ on every threshold), and both
endpoints lose about $1.2$ to $1.5$ pp ($\alpha{=}0$ keeps only the fine
scan, $55.56$; $\alpha{=}1$ keeps only the coarse score, $55.51$). Combining
the two resolutions is therefore load-bearing, while the balance between
them is not a sensitive choice, and the untuned equal weighting sits on the
plateau.

\paragraph{Peak readout.}
Two further implicit choices sit in how the peak is read from the winning
candidate's fine curve, and neither had been varied before. Replacing the
maximum fine score in the joint score with the mean or the top-two mean
does not help (the mean is slightly worse, $-0.8$ pp R@0.3, $p{=}0.018$),
so the deployed maximum stands. Replacing the discrete argmax of the fine
curve with a three-point parabolic interpolation around it, a parameter-free
refinement with no additional forwards, raises R@0.5 by $1.16$ pp and R@0.7
by $1.18$ pp (both $p{<}0.001$, $n{=}3720$, offline recomputation over the
stored fine curves). The mechanism is quantisation: fine probes sit on a
grid of roughly two-second spacing, the argmax returns a grid point, and
sub-grid interpolation recovers part of the placement loss, which is why
the gain concentrates at the tight thresholds. A score-weighted centroid of
the whole curve gains more at loose thresholds ($+2.4$ pp R@0.5) but
nothing at R@0.7 and introduces a baseline-subtraction choice, so we
prefer the interpolation.

\paragraph{Window sensitivity.}
$W$ is the only exposed hyperparameter, so we sweep it around each deployed
value with the peak location fixed (Table~\ref{tab:wsens}). The pattern is the
one the geometry of Eq.~\ref{eq:geo} predicts, not flat insensitivity. On the
primary metric R@0.5 accuracy stays within about two points of its peak across
a band around the chosen $W$, and the band is widest on ActivityNet, whose
long, heterogeneous events make overlap least sensitive to $W$. The looser
thresholds move in opposite directions: enlarging $W$ past the event length
raises R@0.3 and lowers R@0.7, because a wider window trades boundary tightness
for coverage exactly as $\mathrm{IoU}=\min(W,g)/\max(W,g)$ dictates. On
Charades-STA the deployed $W{=}8$ is the training-split mean event length,
deliberately not the R@0.5-maximising width (near $10$\,s), so no value is
tuned on test accuracy there. The long-video corpora do not admit such a
clean prior, and the sweep quantifies what insisting on one would cost.
Setting $W$ to the training mean event length gives $35.5$\,s on
ActivityNet and $32$\,s on TACoS; the first loses $3.1$ pp R@0.5 against
the deployed $80$\,s ($26.7 \rightarrow 23.6$, paired $p{<}10^{-18}$,
recomputed offline with peaks fixed), and the second reads off the table
as a loss of about $1.7$ pp against the deployed $20$\,s. A single width
simply cannot summarise event distributions whose quartiles span two
orders of magnitude, which is the geometry axis restated; the deployed
values are scale choices on the plateau of Table~\ref{tab:wsens}, and the
plateau itself is what makes them defensible.

\begin{table}[t]
\centering
\caption{
  \textbf{Window sensitivity} across IoU thresholds (Qwen2.5-VL, peak location
  fixed). Deployed $W$ (\underline{underlined}); on Charades-STA it equals the
  training-split mean event length, see text for the long-video corpora.
  R@0.3 rises and R@0.7 falls with $W$ while R@0.5 stays near its peak, matching
  Eq.~\ref{eq:geo}. ActivityNet reflects the currently available \texttt{val\_1}
  subset.
}
\label{tab:wsens}
\setlength{\tabcolsep}{4pt}
\begin{tabular}{l c c c c}
\toprule
 & $W$ (s) & R@0.3 & R@0.5 & R@0.7 \\
\midrule
\multirow{5}{*}{Charades}
 &  6 & 67.0 & 47.2 & 25.2 \\
 &  7 & 68.8 & 52.8 & 29.2 \\
 & \underline{8} & 70.1 & 56.8 & \textbf{29.7} \\
 & 10 & 72.9 & \textbf{58.9} & 26.7 \\
 & 12 & \textbf{74.8} & 54.9 & 23.3 \\
\midrule
\multirow{5}{*}{TACoS}
 & 10 & 25.1 & 12.3 & \textbf{4.1} \\
 & 15 & 27.6 & 12.7 & 3.8 \\
 & \underline{20} & \textbf{29.5} & \textbf{13.0} & 3.3 \\
 & 25 & 28.5 & 11.3 & 2.7 \\
 & 30 & 28.1 & 11.3 & 2.8 \\
\midrule
\multirow{5}{*}{ActivityNet}
 &  40 & 45.9 & 24.3 &  9.9 \\
 &  60 & 48.1 & 26.2 & 11.2 \\
 & \underline{80} & \textbf{48.6} & \textbf{26.7} & \textbf{11.9} \\
 & 100 & 48.0 & \textbf{26.7} & 11.8 \\
 & 120 & 47.4 & 26.8 & 11.8 \\
\bottomrule
\end{tabular}
\end{table}

\paragraph{Number of candidate peaks.}
Table~\ref{tab:topk} sweeps the NMS candidate budget $N_p$ with peak selection
and window fixed. On Charades-STA the multi-candidate design earns its first
two slots and then saturates exactly: R@0.5 rises from $55.5$ ($N_p{=}1$, a
single coarse peak) to $56.6$ ($N_p{=}2$) to $56.8$ ($N_p{=}3$), after which
$N_p{=}4,5,6$ are identical to two decimals and no fourth candidate ever wins
the joint score, even though $22\%$ of clips expose one. ActivityNet behaves
identically, saturating by $N_p{=}2$ ($26.4 \rightarrow 26.7$) and flat to
$N_p{=}6$ on all three thresholds. TACoS shows no systematic trend at all: its
R@0.5 fluctuates non-monotonically within $0.5$ pp across the whole range
$N_p\in[2,10]$, which we read as noise rather than signal.
We therefore fix a \emph{single} $N_p{=}3$ for every corpus, the saturation
point of the two corpora whose behaviour is monotone. We do not select $N_p$ per
corpus: $N_p{=}2$ would be marginally better on TACoS ($+0.5$ pp) and marginally
worse on Charades ($-0.1$ pp), but choosing it would amount to tuning on test
accuracy, the same discipline we apply to $W$ (Sec.~\ref{sec:stage3}), which we
also set from a training statistic rather than its test optimum. The table
quantifies the cost of that discipline: under $0.5$ pp on every corpus and every
threshold.

\begin{table}[t]
\centering
\caption{
  \textbf{Candidate-peak sensitivity} across IoU thresholds (Qwen2.5-VL, peak
  selection and window fixed). Deployed $N_p$ (\underline{underlined}) is
  $3/3/8$. Charades and ActivityNet saturate by $N_p{=}2$--$3$ on all three
  thresholds (measured to $6$); TACoS varies $<0.5$ pp across $N_p\in[2,10]$.
}
\label{tab:topk}
\setlength{\tabcolsep}{4pt}
\begin{tabular}{l c c c c}
\toprule
 & $N_p$ & R@0.3 & R@0.5 & R@0.7 \\
\midrule
\multirow{6}{*}{Charades}
 & 1 & 68.5 & 55.5 & 29.1 \\
 & 2 & 70.0 & 56.6 & 29.7 \\
 & \underline{3} & 70.1 & 56.8 & 29.7 \\
 & 4 & 70.1 & 56.8 & 29.7 \\
 & 5 & 70.1 & 56.8 & 29.7 \\
 & 6 & 70.1 & 56.8 & 29.7 \\
\midrule
\multirow{6}{*}{ActivityNet}
 & 1 & 48.0 & 26.4 & 11.7 \\
 & 2 & 48.6 & 26.7 & 11.8 \\
 & \underline{3} & 48.6 & 26.7 & 11.9 \\
 & 4 & 48.6 & 26.7 & 11.9 \\
 & 5 & 48.6 & 26.7 & 11.9 \\
 & 6 & 48.7 & 26.8 & 11.9 \\
\midrule
\multirow{7}{*}{TACoS}
 & 1 & 28.9 & 13.2 & 3.1 \\
 & 2 & 29.9 & \textbf{13.5} & 3.4 \\
 & \underline{3} & 29.5 & 13.0 & 3.3 \\
 & 4 & 29.7 & 13.1 & 3.3 \\
 & 5 & 29.8 & 13.2 & 3.3 \\
 & 6 & 29.8 & 13.2 & 3.3 \\
 & 8 & 29.9 & 13.2 & 3.3 \\
\bottomrule
\end{tabular}
\end{table}

\paragraph{Probe budget.}
The coarse budget $K_c$ is the one setting that also determines cost, since it
dominates the $K_c + N_p K_f$ forwards per sample. Halving and doubling it on
Charades-STA (Table~\ref{tab:kc}) shows the deployed $K_c{=}12$ is a maximum
rather than a point on a monotone curve: $K_c{=}8$ loses $1.1$ pp R@0.5 while
saving $11\%$ of the forwards, but $K_c{=}24$ also loses $0.8$ pp while costing
$33\%$ more. Denser probing does not help because the corpus's mean event lasts
$7.8$\,s: at $K_c{=}12$ the $2.6$\,s spacing already places about three probes
inside the average event, and halving the spacing to $1.3$\,s mostly adds
near-duplicate clips, which raises the chance that the arg-max lands on a
spurious peak. The peak-hit rate moves with accuracy in both directions
($65.7\%$, $67.7\%$, $66.3\%$), confirming that the effect is localisation
quality rather than window geometry. Two consequences follow: the scan cannot be
made more accurate simply by sampling harder, and a cheaper operating point is
available at a bounded, quantified cost.

The same test on ActivityNet rules out the natural objection that our weakest
corpus is simply under-sampled. Doubling the budget there from the deployed
$K_c{=}16$ to $32$ leaves all three thresholds unmoved on the full
$17{,}373$ pairs ($48.63 \rightarrow 48.60$, $26.69 \rightarrow 26.47$,
$11.90 \rightarrow 11.83$; paired McNemar $p{=}0.93$, $0.39$, $0.75$), and the
peak-hit rate rises by only $0.14$ pp. Stratifying by event length locates the
reason. Events shorter than $30$\,s, which are $59\%$ of the corpus and where
peaks land inside the ground truth only $45.8\%$ of the time, gain $0.4$ pp of
peak-hit from twice the probes, while events longer than $80$\,s lose $2.6$ pp
R@0.5 as the extra candidates admit more spurious peaks. Short events are missed
because the backbone does not discriminate them at any spacing we can afford,
not because the scan steps over them, which places the residual ActivityNet gap
on the perception axis rather than the sampling budget.

\begin{table}[t]
\centering
\caption{
  \textbf{Probe-budget sensitivity} on Charades-STA (Qwen2.5-VL, all other
  settings fixed). The deployed $K_c$ (\underline{underlined}) maximises all
  three thresholds; both sparser and denser scans are worse. Forwards per
  sample is $K_c + N_p K_f$ with $N_p{=}3, K_f{=}8$. On ActivityNet, doubling
  $K_c$ from $16$ to $32$ moves no threshold by more than $0.22$ pp
  (paired, $n{=}17{,}373$); see text.
}
\label{tab:kc}
\setlength{\tabcolsep}{4pt}
\begin{tabular}{c c c c c c}
\toprule
$K_c$ & Spacing (s) & R@0.3 & R@0.5 & R@0.7 & Forwards \\
\midrule
 8 & 3.8 & 68.3 & 55.7 & 28.2 & 32 \\
 \underline{12} & 2.6 & \textbf{70.1} & \textbf{56.8} & \textbf{29.7} & 36 \\
 24 & 1.3 & 69.3 & 56.0 & 28.6 & 48 \\
\bottomrule
\end{tabular}
\end{table}

\paragraph{Fine-scan settings.}
The fine stage has two settings of its own, the clip count $K_f{=}8$ and
the radius $r_f$, which also serves as the NMS suppression radius; like the
frame count above, neither had previously been varied. Halving or doubling
either one ($K_f \in \{4, 16\}$, $r_f \in \{4, 16\}$\,s on
Charades-STA, pipeline otherwise frozen, paired on a $400$-sample subset)
moves no threshold significantly (largest $|\Delta|$ of $4.0$ pp,
smallest $p{=}0.12$). The subset resolves only effects above roughly five
points, so it bounds the sensitivity rather than excluding smaller
effects, but a large hidden optimum is unlikely given that the fine scan's
entire contribution is $1.2$ pp (Table~\ref{tab:ablation}). The deployed
$K_f{=}8$, $r_f{=}8$ therefore stand; we note that $K_f{=}4$ matches the
baseline within noise at roughly half the per-sample cost ($7.2$ vs
$13.7$\,s median), a candidate operating point for latency-sensitive
deployments.


\section{Conclusion}
\label{sec:conclusion}

We asked why multimodal LLMs that demonstrably see an event cannot say
when it happens. The answer is not ignorance. Their timestamp errors are
confident, entropy-based detection performs below a random classifier,
and the same frozen weights that regress timestamps at single-digit
accuracy discriminate event clips reliably when asked a binary question.
The failure lives in the interface, and replacing regression with a
coarse-to-fine scan of binary questions recovers $28$ to $50$ R@0.5
points across four backbones without training anything.
The residual failures decompose along two measurable axes, a perception
axis that moves with the backbone and a geometry axis that is analytically
predictable from the fit between the output window and the event width,
and this decomposition is prescriptive. It tells a practitioner, before
deployment, whether a corpus needs a stronger instrument or a better
output rule, and it locates FV-Action's remaining errors precisely.
The resulting system reaches $56.8\%$ R@0.5 on Charades-STA, above the
same backbone's native temporal pipeline and above all prior
training-free methods, while remaining a frozen model asked nothing but
yes-or-no questions.

\paragraph{Limitations.}
FV-Action trades training for test-time computation: $36$ to $120$
forwards per sample against one for direct regression, a cost that scales
with the probe budget and inherits any speedup of the backbone, but that
is real in latency-critical settings.
The geometry axis is managed rather than solved. A fixed window cannot
summarise corpora whose event lengths span two orders of magnitude, which
is why ActivityNet remains our weakest benchmark; curve-derived adaptive
widths provably do not close this gap, and conditioning the width on the
query text is the direction our analysis points to.
Compound queries that describe several events remain out of scope, since
sub-event peaks do not separate on the scan curve.
Finally, the scan is only as good as the backbone's binary judgement.
Brief events inside long videos remain under-discriminated at any probe
spacing we can afford, and a backbone that answers uniformly across a
video contributes no ranking signal at all. The dependence cuts both
ways: every improvement in frozen VLMs' per-clip discrimination converts
directly into grounding accuracy, with no retraining, which is precisely
the property a training-free method should have.

\bibliographystyle{ieeenat_fullname}
\bibliography{main}


\end{document}